\documentclass{article}

\PassOptionsToPackage{numbers}{natbib}

\usepackage[preprint]{neurips_2023}

\usepackage[T1]{fontenc}    
\usepackage{hyperref}       
\usepackage{url}            
\usepackage{booktabs}       
\usepackage{amsfonts}       
\usepackage{nicefrac}       
\usepackage{microtype}      
\usepackage{xcolor}         
\usepackage{soul}
\usepackage{graphicx}
\usepackage{verbatim}
\usepackage{caption}
\usepackage{subcaption}
\usepackage{footmisc}
\usepackage{booktabs}
\usepackage{hhline}
\usepackage{multirow}
\usepackage{arydshln}
\usepackage{float}
\usepackage{xltabular}
\usepackage{lmodern}
\usepackage[most]{tcolorbox}
\tcbuselibrary{skins}
\usepackage{algpseudocode}
\usepackage{longtable}
\usepackage{framed}
\usepackage{hyperref}
\usepackage{enumitem}  
\usepackage{adjustbox}
\usepackage{makecell}
\usepackage{multirow}
\usepackage{booktabs}
\usepackage{pdflscape}
\usepackage{listings}
\usepackage{placeins}

\tcbset{
  aibox/.style={
    width=\textwidth,
    top=0pt, bottom=0pt, left=5pt, right=5pt,
    colback=white,
    colframe=black,
    colbacktitle=black,
    enhanced,
    center,
    attach boxed title to top left={yshift=-0.1in,xshift=0.15in},
    boxed title style={boxrule=0pt,colframe=white,},
  }
}
\newtcolorbox{AIbox}[2][]{aibox,title=#2,#1}

\usepackage[ruled,vlined]{algorithm2e}
\newcommand{\squishlist}{
   \begin{list}{$\bullet$}
    { \setlength{\itemsep}{0pt}      \setlength{\parsep}{3pt}
      \setlength{\topsep}{3pt}       \setlength{\partopsep}{0pt}
      \setlength{\leftmargin}{1.5em} \setlength{\labelwidth}{1em}
      \setlength{\labelsep}{0.5em} } }

\newcommand{\squishlisttwo}{
   \begin{list}{$\bullet$}
    { \setlength{\itemsep}{0pt}    \setlength{\parsep}{0pt}
      \setlength{\topsep}{0pt}     \setlength{\partopsep}{0pt}
      \setlength{\leftmargin}{2em} \setlength{\labelwidth}{1.5em}
      \setlength{\labelsep}{0.5em} } }

\newcommand{\squishend}{
    \end{list}  }

\usepackage{blindtext}

\newtcolorbox[list inside=mybox,auto counter,number within=section]{MyBox}{colbacktitle=yellow,coltitle=black,title={MyBox \thetcbcounter}}

\newcommand{\sysname}{LoRA Land}

\usepackage{tikz}

\title{%
\raisebox{-0.3cm}{\includegraphics[width=1cm, height=1cm]{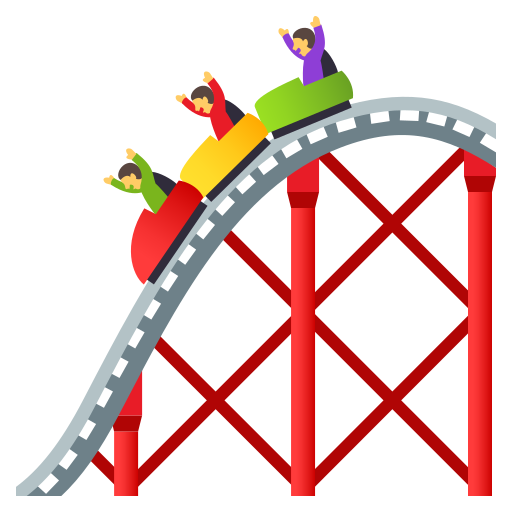}}\ {\sysname: 310 Fine-tuned LLMs that Rival GPT-4, A Technical Report}
}

\author{\textbf{Justin Zhao, Timothy Wang} \\
\textbf{Wael Abid, Geoffrey Angus, Arnav Garg, Jeffery Kinnison, Alex Sherstinsky}, \\
\textbf{Piero Molino, Travis Addair, Devvret Rishi} \\
\AND Predibase\vspace{0em}}

\begin{document}

\maketitle
\begin{abstract}

Low Rank Adaptation (LoRA) has emerged as one of the most widely adopted methods for Parameter Efficient Fine-Tuning (PEFT) of Large Language Models (LLMs). LoRA reduces the number of trainable parameters and memory usage while achieving comparable performance to full fine-tuning. We aim to assess the viability of training and serving LLMs fine-tuned with LoRA in real-world applications. First, we measure the quality of LLMs fine-tuned with quantized low rank adapters across 10 base models and 31 tasks for a total of 310 models. We find that 4-bit LoRA fine-tuned models outperform base models by 34 points and GPT-4 by 10 points on average. Second, we investigate the most effective base models for fine-tuning and assess the correlative and predictive capacities of task complexity heuristics in forecasting the outcomes of fine-tuning. Finally, we evaluate the latency and concurrency capabilities of LoRAX, an open-source Multi-LoRA inference server that facilitates the deployment of multiple LoRA fine-tuned models on a single GPU using shared base model weights and dynamic adapter loading. LoRAX powers \href{https://predibase.com/lora-land}{LoRA Land}, a web application that hosts 25 LoRA fine-tuned Mistral-7B LLMs on a single NVIDIA A$100$ GPU with $80$GB memory. LoRA Land highlights the quality and cost-effectiveness of employing multiple specialized LLMs over a single, general-purpose LLM.

\begin{figure}[ht]
    \centering
    \scalebox{0.9}{
    \includegraphics[width=\textwidth]{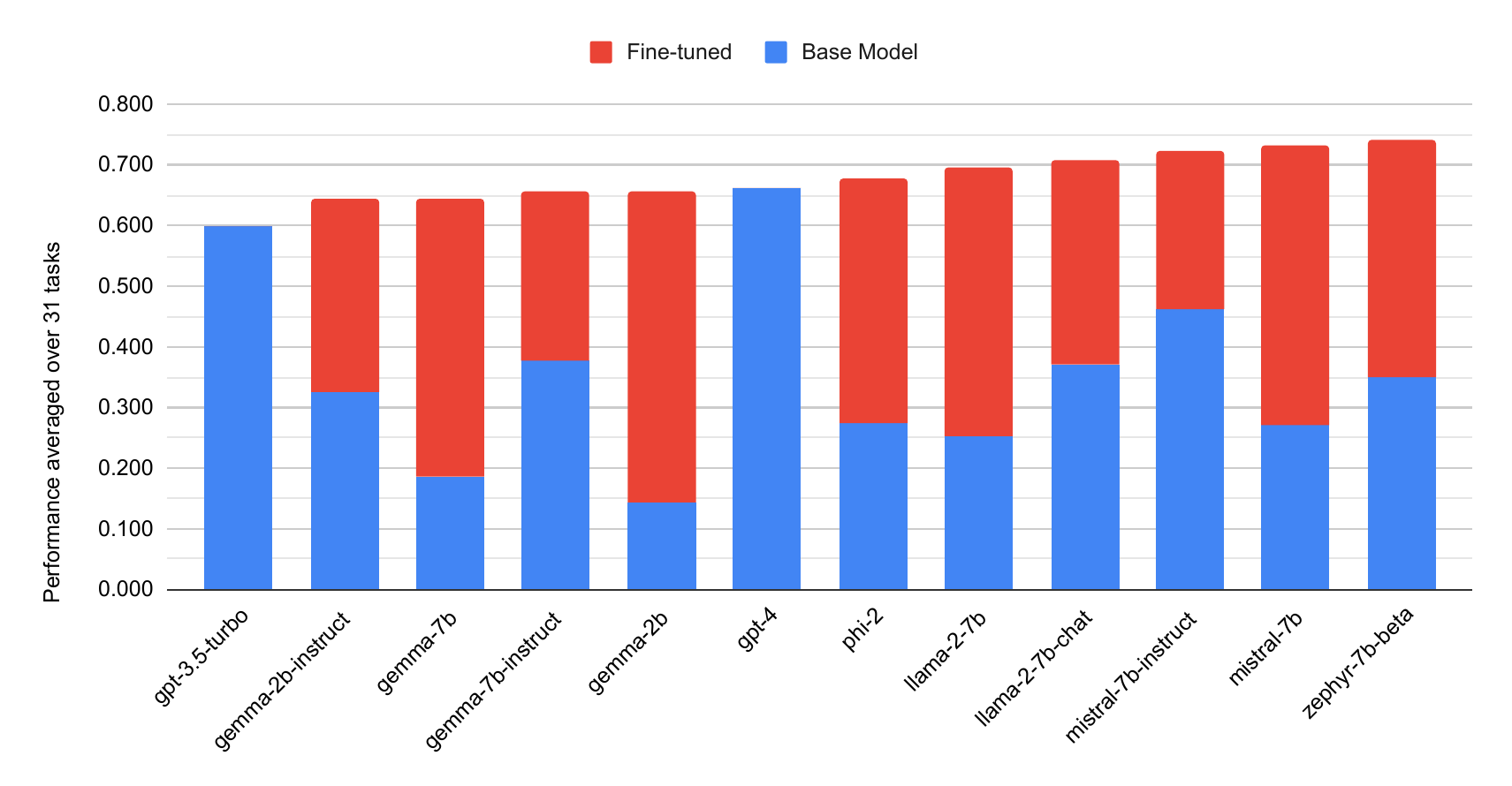}
    }
    \caption{Average model performance for GPT-3.5, GPT-4, and 310 LLMs, before and after fine-tuning with LoRA, across 31 different tasks and 10 different base models. Zephyr-7b and Mistral-7b models exhibit the best performance after LoRA-based fine-tuning.}
    \label{fig:orca2}
\end{figure}

\end{abstract}

\section{Introduction}

Fine-tuning Large Language Models (LLMs)~\cite{67minaee2024survey, 52stanfordOpportunitiesRisks} is a highly effective way to improve their performance, and add desirable or remove undesirable behaviors~\cite{2ouyang2022training, 13houlsby2019peft, 16howard2018finetuning, 17peters2019finetuning}. Low Rank Adaptation (LoRA)~\cite{1hu2021lora} is one of the most widely adopted methods for fine-tuning LLMs, showing significant promise for enabling smaller, specialized models to outperform larger, more general models on specific tasks, with a fraction of trainable parameters, challenging the notion that bigger general models always outperform smaller ones.

Despite the rapid advancement and release of new base models, such as Gemma~\cite{3gemma}, Llama~\cite{4llama}, and Mistral~\cite{5mistral}, which claim ease of fine-tuning across various tasks, comprehensive evaluations of these models remain scarce. Broad knowledge and reasoning-based benchmarks like MMLU~\cite{28hendrycks2020mmlu} and HellaSwag~\cite{29zellers2019zellers} are commonly used in leaderboards like the Open LLM Leaderboard~\cite{34open-llm-leaderboard}, however, this is not necessarily representative of task-specific performance, before or after fine-tuning. Technical reports~\cite{3gemma, 4llama, 5mistral, 35gpt4, 32gemini} often leave training configurations unspecified, with claims of ease of fine-tuning left unmeasured. While the effectiveness of fine-tuning has been broadly demonstrated~\cite{9kohut2023finetuning, 10zheng2024}, the lack of large-scale experimentation leaves several pivotal questions unanswered, particularly regarding the consistency and predictability of performance improvements through fine-tuning, and the impact of model size, base model, and task complexity.

Evaluations are sensitive to prompting, and there are significant variation in the formulations used in publications and libraries\footnote{https://github.com/openai/simple-evals}. Technical reports often showcase model performance using specialized, dataset-specific prompting strategies such as role-playing prompts (e.g. \textit{"Assume you are an expert"}), \textit{maj@k} voting~\cite{48wangselfconsistency}, varied n-shot~\cite{37song2022evalsurvey}, MedPrompt~\cite{33nori2023generalistvsfinetuned}, and chain-of-thought~\cite{36wei2022cot} prompting. While these methods are intended to highlight the optimal capabilities of models, the use of such diverse prompting techniques can make direct comparisons across models and tasks challenging.

In this work, we seek to bridge these gaps by conducting an extensive analysis of LoRA-based fine-tuning across 10 base models and 31 tasks, totaling 310 LLMs fine-tuned with LoRA. We deliberately maintain that all LLMs are fine-tuned with the same training parameters and emphasize querying with zero or single-shot, completion-style prompts, with simple instructions like \textit{"Solve the following multiple choice problem"}. Altogether, this provides a standardized framework to compare and assess the intrinsic capabilities of different base models when fine-tuned with LoRA under consistent conditions, across specific tasks.

We also aim to explore the viability of serving multiple LoRA models in a real-world production application. LoRAX ~\cite{12predibaseLoRAExchange} enables serving multiple LoRA models simultaneously on a single GPU by leveraging shared base model weights and dynamic adapter loading [12]. We measure latency and concurrency metrics of this library. We use LoRAX to deploy 25 fine-tuned LLM served on a single A100\footnote{https://www.nvidia.com/en-us/data-center/a100/} in the LoRA Land web application. Our successful implementation showcases the economic efficiency of serving multiple LoRA-adapted LLMs for specialized tasks.

Finally, we release all 25 of the fine-tuned models on the LoRA Land web application and their training recipes on (\href{https://huggingface.co/predibase}{Hugging Face}) to allow further analysis and replication by the community.

\section{Related work}

\textbf{Parameter-Efficient Fine-Tuning (PEFT)} methods are designed to reduce the high expense of fine-tuning large-scale models. They achieve this by training a relatively small subset of parameters, compared to the total number of parameters, for adapting to downstream tasks. Existing PEFT strategies can be divided into two categories: \textit{Prompt-based} methods add extra soft tokens (prompts) to the initial input and focus solely on fine-tuning these trainable vectors~\cite{26lester2021prompttuning, Razdaibiedina2023progressiveprompts, wang2021learningtoprompt}. \textit{Adapter-based methods} introduce additional trainable modules into the original frozen backbone~\cite{13houlsby2019peft, 22rebuffi2017residualadapters, 23pfeiffer2020adapterfusion, 24ruckle2020adapterdrop}. LoRA~\cite{1hu2021lora} expands upon adapter-based fine-tuning by adding a small number of trainable low-rank matrices alongside layers of frozen weights, which introduces a negligible inference overhead. Variants of LoRA include works like~\cite{meng2024periodiclora}, which employs SVD decomposition to prune less significant singular values for more efficient updates. Another variation, DoRA ~\cite{dora}, decomposes pre-trained weights into magnitude and direction components while applying LoRA the latter. QLoRA~\cite{11dettmers2023qlora} optimizes LoRA’s design one step further, using 4-bit NF4 weights, double quantization to reduce the memory footprint, and paged optimizers to alleviate memory spikes. In our experiments, we focus on the original implementation of LoRA with 4-bit quantization.

\textbf{Efficient serving of LoRA models.} The main challenges for serving multiple fine-tuned models efficiently are:

\begin{enumerate}
\item \textbf{Scalability:} As the demand for model inference grows, the system must scale efficiently to handle the increased load. This involves not just scaling up the computational resources but also managing the load distribution among models to maintain performance.

\item \textbf{Cost:} The computational resources required to serve multiple fine-tuned models can lead to significant costs. Efficiently managing these costs while maintaining high performance and availability is a major challenge.

\end{enumerate}

Techniques like Segmented Gather Matrix-Vector Multiplication (SGMV)~\cite{38chen2023punica} aim to address these challenges by optimizing the way computations are performed and resources are used. Open source tools like DeepSpeed\footnote{https://github.com/microsoft/DeepSpeed}, FasterTransformer\footnote{https://github.com/NVIDIA/FasterTransformer}, and vLLM~\cite{41kwon2023efficient} also aim to enable cost-effective and scalable serving of fine-tuned models. In this paper, we use LoRAX\footnote{https://github.com/predibase/lorax}, which is specifically designed for the efficient serving of LLMs fine-tuned with LoRA. LoRAX supports dynamic adapter loading so adapters can be downloaded asynchronously during inference, multiple model families like Llama~\cite{4llama} and Mistral~\cite{5mistral}, and bitsandbytes\footnote{https://github.com/TimDettmers/bitsandbytes}-quantized models.

\section{Methodology}

\subsection{Task selection}

In selecting datasets and tasks for our study, we prioritize those that are widely accessible via Kaggle\footnote{https://www.kaggle.com} and HuggingFace\footnote{https://huggingface.co} and those that are commonly used for benchmarking such as those on the Open LLM Leaderboard~\cite{34open-llm-leaderboard}.

Our selection includes datasets like MMLU~\cite{28hendrycks2020mmlu} for broad domain knowledge, Jigsaw~\cite{jigsaw-unintended-bias-in-toxicity-classification} for content moderation, WikiSQL~\cite{zhongSeq2SQL2017} for SQL generation, and GLUE benchmarks~\cite{58wang2018glue}. We categorize the tasks encompassed by these datasets into 5 types:

\begin{itemize}
    \item \textbf{Classic NLP:} Tasks derived from common NLP datasets published between 2018 and 2022 covering tasks like named entity recognition, data-to-text, and headline generation.
    \item \textbf{Coding:} SQL query generation, and Python programming questions, which are mostly centered on algorithms and object-oriented design.
    \item \textbf{Knowledge:} Knowledge-based multiple choice questions.
    \item \textbf{Reasoning:} Reasoning-based multiple choice questions.
    \item \textbf{Math:} Numerical, math-based word problems.
\end{itemize}

\begin{landscape}

\begin{table*}[ht]
\centering

\scalebox{0.5}{

\begin{tabular}{ccllccccccc}
\multirow{2}{*}{\textbf{Category}} & \multirow{2}{*}{\textbf{Task Name}} & \multicolumn{1}{c}{\multirow{2}{*}{\textbf{Task Description}}}        & \multicolumn{1}{c}{\multirow{2}{*}{\textbf{Dataset Link}}}   & \multirow{2}{*}{\textbf{Metric}} & \multirow{2}{*}{\textbf{Range \# Tokens}} & \multirow{2}{*}{\textbf{P95 \# Tokens}} & \multicolumn{3}{c}{\textbf{\# examples}}             & \multicolumn{1}{l}{\multirow{2}{*}{\textbf{Split Used for Evaluation}}} \\
\cmidrule(l{3pt}r{3pt}){8-10}
                                   &                                     & \multicolumn{1}{c}{}                                                  & \multicolumn{1}{c}{}                                         &                                  &                                           &                                         & \textbf{train} & \textbf{validation} & \textbf{test} & \multicolumn{1}{l}{}                                                    \\
\midrule
\multirow{7}{*}{Classic NLP}       & bc5cdr                              & Chemical and disease recognition                                      & hf://tner/bc5cdr                                             & rouge                            & 143 - 570                                 & 226                                     & 5228           & 5330                & 5865          & validation                                                              \\
                                   & conllpp                             & Named entity recognition                                              & hf://conllpp                                                 & rouge                            & 110 - 401                                 & 170                                     & 14041          & 3250                & 3453          & test                                                                    \\
                                   & e2e\_nlg                            & Translation from meaning representation to natural language           & hf://e2e\_nlg                                                & rouge                            & 92 - 213                                  & 153                                     & 42061          & 4672                & 4693          & test                                                                    \\
                                   & tldr\_content\_gen                  & Content generation given a headline                                   & hf://JulesBelveze/tldr\_news                                 & rouge                            & 46 - 425                                  & 204                                     & 7138           & --                  & 794           & test                                                                    \\
                                   & tldr\_headline\_gen                 & Headline generation given news content                                & hf://JulesBelveze/tldr\_news                                 & rouge                            & 41 - 420                                  & 199                                     & 7138           & --                  & 794           & test                                                                    \\
                                   & viggo                               & Translation of video game meaning representations to natural language & hf://GEM/viggo                                               & rouge                            & 151 - 304                                 & 240                                     & 5103           & 714                 & 1083          & test                                                                    \\
                                   & webnlg                              & Translation of triples to natural language                            & hf://web\_nlg (release\_v3.0\_en)                            & rouge                            & 88 - 345                                  & 215                                     & 13211          & 1667                & 5713          & test                                                                    \\
\midrule
\multirow{2}{*}{Coding}            & magicoder                           & Coding tasks in multiple languages                                    & hf://ise-uiuc/Magicoder-OSS-Instruct-75K                     & humaneval                        & 141 - 1661                                & 805                                     & 75197          & --                  & --            & (human\_eval)                                                           \\
                                   & wikisql                             & SQL generation given a table and question                             & hf://wikisql                                                 & rouge                            & 198 - 72472                               & 1941                                    & 56355          & 8421                & 15878         & test                                                                    \\
\midrule
\multirow{8}{*}{Knowledge}         & boolq                               & Knowledge-based yes/no questions.                                     & hf://google/boolq                                            & accuracy                         & 30 - 898                                  & 271                                     & 9427           & 3270                & --            & validation                                                              \\
                                   & dbpedia                             & Topic extraction from a news article and title                        & hf://fancyzhx/dbpedia\_14                                    & accuracy                         & 102 - 387                                 & 211                                     & 560000         & --                  & 70000         & test                                                                    \\
                                   & customer\_support                   & Customer support call classification given call transcript            & github://cricketclub/gridspace-stanford-harper-valley        & accuracy                         & 151 - 679                                 & 377                                     & 1245           & 245                 & 391           & test                                                                    \\
                                   & glue\_qnli                          & Does the response answer the question?                                & hf://glue/viewer/qnli                                        & accuracy                         & 52 - 350                                  & 123                                     & 104743         & 5463                & 5463          & validation                                                              \\
                                   & glue\_stsb                          & How similar are the sentences?                                        & hf://glue/viewer/stsb                                        & mae                              & 74 - 187                                  & 124                                     & 5749           & 1500                & 1379          & validation                                                              \\
                                   & legal                               & Legal document classification                                         & kaggle://bahushruth/legalclausedataset                       & rouge                            & 143 - 885                                 & 489                                     & 17000          & 2000                & 1000          & test                                                                    \\
                                   & reuters                             & Topic extraction from Reuters news articles                           & hf://reuters21578/viewer/ModLewis (modlewis)                 & rouge                            & 51 - 2056                                 & 637                                     & 13625          & --                  & 6188          & test                                                                    \\
                                   & mmlu                                & General domain multiple-choice questions                              & hf://cais/mmlu/viewer/all                                    & accuracy                         & 47 - 1491                                 & 578                                     & 99842          & 1531                & 14042         & validation                                                              \\
\midrule
\multirow{13}{*}{Reasoning}        & winogrande                          & Common sense 2-option task                                            & hf://winogrande                                              & accuracy                         & 48 - 75                                   & 63                                      & 9248           & 1767                & 1267          & test                                                                    \\
                                   & arc\_combined                       & Multiple-choice science questions                                     & hf://allenai/ai2\_arc                                        & accuracy                         & 68 - 232                                  & 143                                     & 3370           & 869                 & 3548          & test                                                                    \\
                                   & glue\_cola                          & Grammar and syntax acceptability                                      & hf://glue/viewer/cola                                        & accuracy                         & 45 - 87                                   & 59                                      & 8551           & 1043                & 1063          & validation                                                              \\
                                   & glue\_mnli                          & Does the hypothesis entail the premise?                               & hf://nyu-mll/glue/viewer/mnli                                & accuracy                         & 64 - 339                                  & 128                                     & 392702         & 19647               & 19643         & validation                                                              \\
                                   & glue\_mrpc                          & Do the sentences have the same meaning?                               & hf://glue/viewer/mrpc                                        & accuracy                         & 67 - 157                                  & 123                                     & 3668           & 408                 & 1725          & validation                                                              \\
                                   & glue\_qqp                           & Do the questions have the same meaning?                               & hf://glue/viewer/qqp                                         & accuracy                         & 60 - 351                                  & 102                                     & 363846         & 40430               & 390965        & validation                                                              \\
                                   & glue\_sst2                          & Binary sentiment detection                                            & hf://glue/viewer/sst2                                        & accuracy                         & 33 - 91                                   & 63                                      & 67349          & 872                 & 1821          & validation                                                              \\
                                   & glue\_wnli                          & Pronoun resolution                                                    & hf://glue/viewer/wnli                                        & accuracy                         & 73 - 160                                  & 134                                     & 635            & 71                  & 146           & validation                                                              \\
                                   & covid                               & Sentiment detection of COVID-19 tweets                                & kaggle://datatattle/covid-19-nlp-text-classification         & accuracy                         & 131 - 292                                 & 223                                     & 37361          & --                  & 3798          & test                                                                    \\
                                   & hellaswag                           & Multiple-choice sentence completion                                   & hf://Rowan/hellaswag                                         & accuracy                         & 120 - 407                                 & 341                                     & 39905          & 10003               & 10042         & validation                                                              \\
                                   & hellaswag\_processed                & Sentence completion                                                   & hf://Rowan/hellaswag                                         & rouge                            & 75 - 205                                  & 185                                     & 39905          & 10003               & 10042         & validation                                                              \\
                                   & jigsaw                              & Toxic comment classification                                          & kaggle://c/jigsaw-unintended-bias-in-toxicity-classification & accuracy                         & 409 - 715                                 & 601                                     & 159571         & --                  & 153164        & test                                                                    \\
                                   & drop                                & Question answering given a passage                                    & hf://drop                                                    & rouge                            & 87 - 2275                                 & 571                                     & 77400          & 9535                & --            & validation                                                              \\
\midrule
Math                               & gsm8k                               & Grade school math problems                                            & hf://gsm8k (main)                                            & accuracy                         & 58 - 465                                  & 276                                     & 7473           & --                  & 1319          & test                                                                   
\end{tabular}

}

\vspace{10mm}
\captionsetup{type=figure}
\caption{Tasks and datasets used. \textit{tldr\_news} and \textit{hellaswag} datasets are used for multiple tasks. The length of the texts vary substantially across tasks. Many tasks and datasets exhibit a long-tail distribution, where a small number of examples have significantly longer sequences than the average. Token counts are based on the tiktoken package \cite{githubGitHubOpenaitiktoken}. \label{table:datasets}}

\end{table*}
\end{landscape}

\subsection{Prompt selection}

\begin{figure}[ht]
    \centering
    \includegraphics[width=\textwidth]{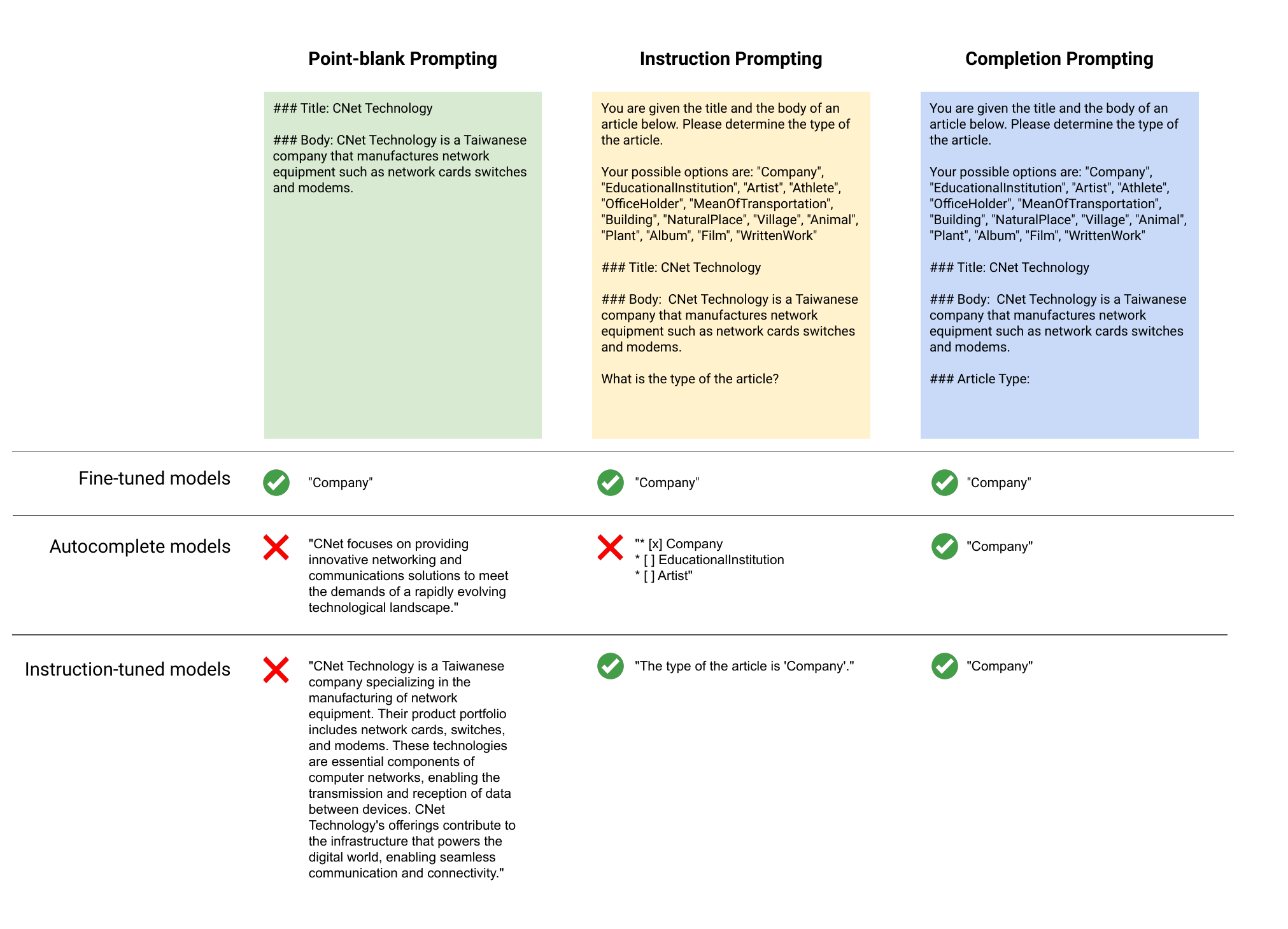}
    \caption{Examples of different styles of prompting. To maintain using the same prompts when comparing models and to ensure the highest likelihood of success amongst all types of models (fine-tuned, auto-complete, or instruction-tuned), all of our prompts adhere to completion style.}
    \label{fig:promptdesign}
\end{figure}

Previous studies have demonstrated the potential of leveraging prompt engineering techniques, such as the use of majority voting [48], the inclusion of multiple in-context examples (n-shot)~\cite{37song2022evalsurvey}, MedPrompt~\cite{33nori2023generalistvsfinetuned}, chain-of-thought prompting~\cite{36wei2022cot}, etc., to enhance model performance on specific tasks. 

In our evaluations, we consciously choose \textbf{not} to employ additional prompt engineering or tuning strategies for any specific dataset, task, or model. Although using more in-context examples or a more selective approach in n-shot prompting might yield better results, we prioritize reproducibility and the minimization of biases that could arise from customized in-context learning. Instead, we opt to use simple zero or single-shot completion-style prompts for all tasks. Our prompts are written in the completion style, described in Figure \ref{fig:promptdesign}, to provide a fair comparison across fine-tuned, instruction-tuned, and auto-complete models. For classification tasks, the prompt includes all possible classes to inform the model's responses. For more specialized tasks, where describing the expected output format is challenging, we use a single in-context example -- the first example from the published training split -- to guide the model.

\begin{table}[ht]

    \centering
    \includegraphics[width=\textwidth]{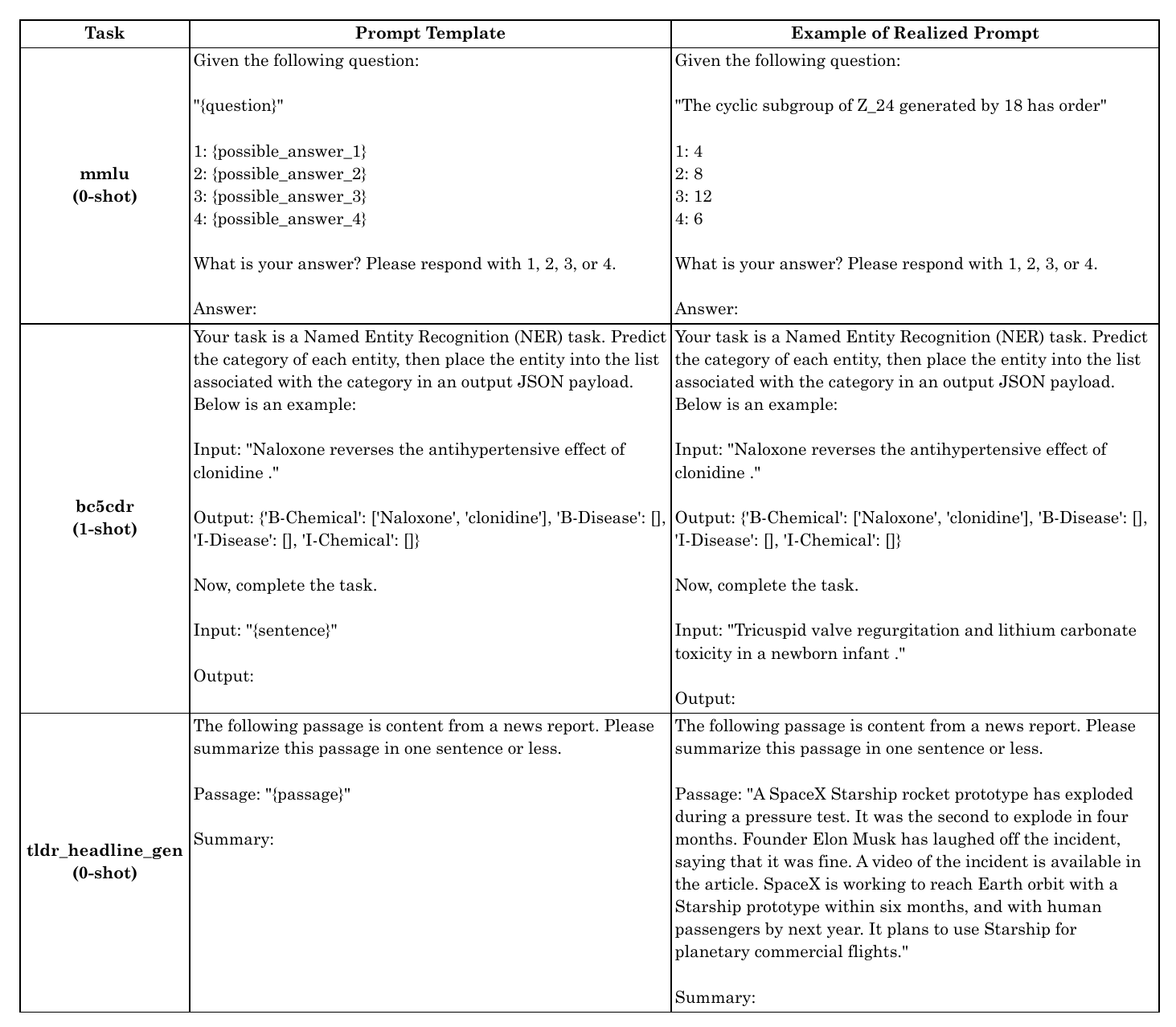}
    \caption{Examples of prompts that are used in this study, all written in completion style. For more specialized tasks, where describing the expected output format is challenging (e.g. \textit{bc5cdr}), we use a single in-context example — the first example from the published training split — to guide the model.}

\end{table}

Finally, we follow prescribed prompt tagging conventions for each model, as outlined in the respective model's documentation on HuggingFace, to ensure proper querying of pre-trained and instruction-tuned base models. This includes using \textbf{"<s>[INST] ... [/INST]"} for prompts intended for Mistral Instruct, and \textbf{"<bos><start\_of\_turn>user ... <end\_of\_turn><start\_of\_turn><model>"} for Gemma's instruction-tuned models.
For detailed information on the exact prompt templates applied to each task and model, please see Appendix \ref{sec:appendixprompts}.

\subsection{Base models}

All base models are listed in Table \ref{table:basemodels}. We use GPT-4 (gpt-4-0613) and GPT-3.5-Turbo (gpt-3.5-turbo-0125) as two strong LLM baselines. Our selection of these ten base models was guided by several key considerations, including their widespread adoption within the AI community, availability with permissive licenses, and availability of technical reports. We specifically choose base models with $\leq 8$ billion parameters to ensure that each model can be efficiently trained within the resource limits of a single A10G GPU. 

\begin{table*}[ht]
\centering  
\footnotesize 
\begin{tabular}{llll}
\textbf{Model Name}      & \textbf{Creator} & \textbf{\# of Parameters} & \textbf{Date Released} \\
\midrule
Llama-2-7b               & Meta             & 7B                        & July 18, 2023          \\
Llama-2-7b-chat          & Meta             & 7B                        & July 18, 2023          \\
Mistral-7b-v0.1          & Mistral AI       & 7.24B                     & September 20, 2023     \\
Mistral-7b-Instruct-v0.1 & Mistral AI       & 7.24B                     & September 27, 2023     \\
Zephyr-7b                & Hugging Face     & 7.24B                     & October 26, 2023       \\
Phi-2b                   & Microsoft        & 2.78B                     & December 13, 2023      \\
Gemma-2b                 & Google           & 2.51B                     & February 21, 2024      \\
Gemma-2b-it              & Google           & 2.51B                     & February 21, 2024      \\
Gemma-7b                 & Google           & 8.54B                     & February 21, 2024      \\
Gemma-7b-it              & Google           & 8.54B                     & February 21, 2024     
\end{tabular}

\caption{Base models used in LoRA-based fine-tuning experiments. To train all models on A10G hardware, all chosen base models are ~7B parameters or smaller.}
\label{table:basemodels}
\end{table*}

\subsection{Training parameters}

Each model is trained with published train splits\footnote{\textit{customer\_support} and \textit{legal} are the only two tasks in our list without official splits. The exact splits for these datasets are published on <\href{github.com/predibase/lora-bakeoff}{github.com/predibase/lora-bakeoff}>.}. Each model is trained for $40000$ training steps with batch size 1, 4-bit quantization using bitsandbytes and a LoRA rank of $8$. We use the paged adam optimizer\cite{11dettmers2023qlora}, a learning rate of $0.002$, and a cosine learning rate scheduler with a $0.03$ warm-up fraction ($1200$ training steps). Gradients are applied over $16$ accumulation steps for an effective batch size of $16$.

These training parameters, combined with gradient checkpointing, allow each LLM to be fine-tuned on a single A10 GPU with 24 GB of memory. For tasks where training on the full sequence lengths would still produce a GPU Out-Of-Memory (OOM) error, we first truncate example inputs to a maximum sequence length set as the 95th percentile of all task inputs.

To guarantee a consistent and straightforward basis of comparison across models, no additional hyperparameter tuning is applied to any specific dataset, task, or base model.

\begin{figure}[ht]
    \centering
    \scalebox{0.5}{
    \includegraphics[width=\textwidth]{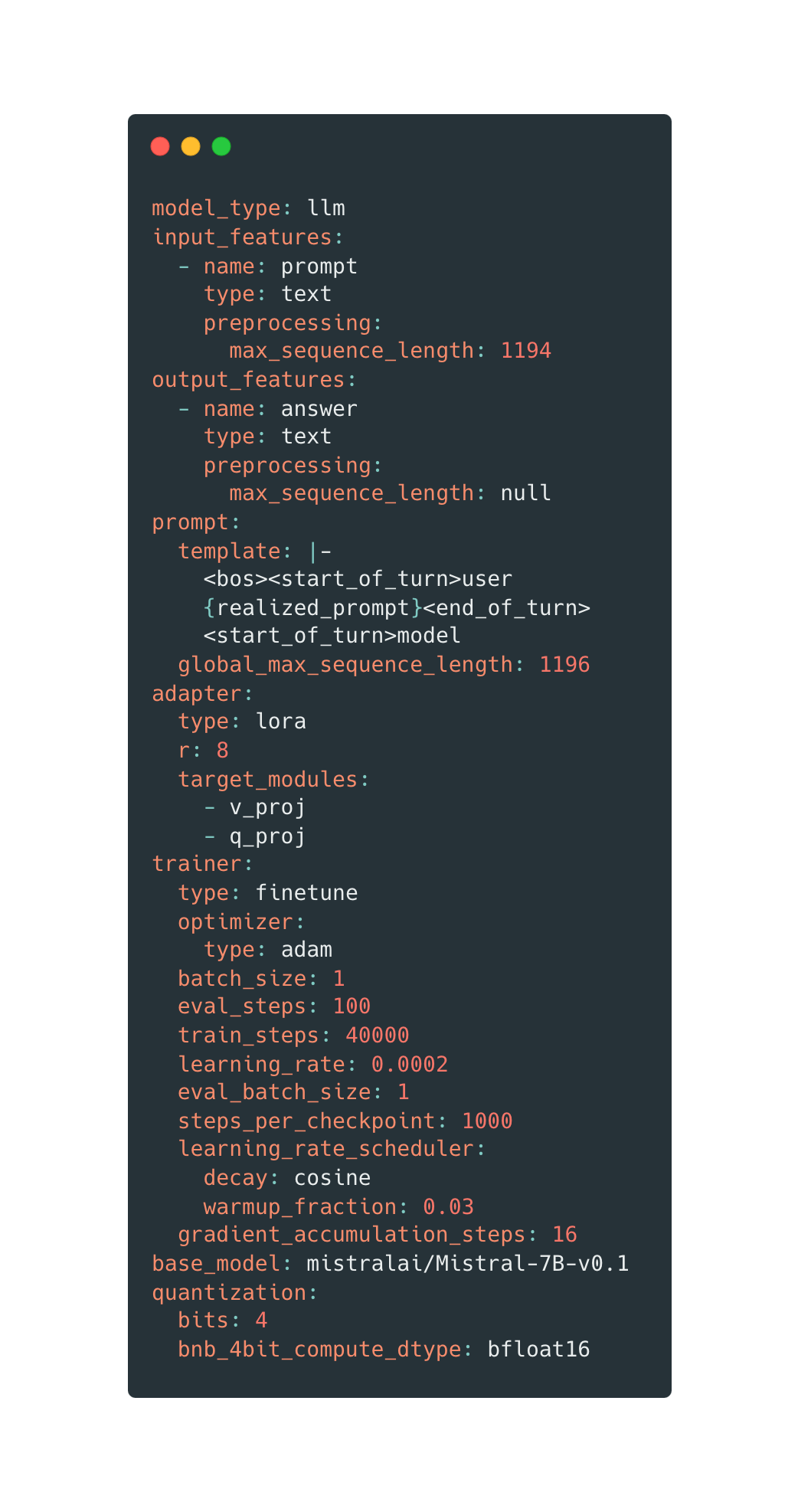}
    }
    \caption{Example LLM model training configuration for LoRA-based fine-tuning. Based on Ludwig \protect\cite{62Molino2019}.}
    \label{fig:sampleconfig}
\end{figure}

Training recipes for each model are provided as Ludwig~\cite{62Molino2019} configurations for each of the fine-tuned LLMs and can be found at \href{https://huggingface.co/predibase}{https://huggingface.co/predibase}. Figure \ref{fig:sampleconfig} shows an example of a config.

\subsection{Evaluation}

As specified in Table \ref{table:datasets}, models are evaluated on the test split if it exists and is labeled, or the validation set otherwise\footnote{MMLU has a published test set with labels, however, we use validation split to be consistent with the HELM benchmark~\cite{liang2023helm}}. We employ a tailored set of evaluation metrics to accurately assess the performance across all of the tasks. We use accuracy for classification tasks, (1 - mean absolute error) for regression tasks\footnote{Mean absolute error (MAE) is used because the range of target values are integer-like and small.}, and rouge-L\footnote{Text generation tasks are complicated to evaluate automatically~\cite{53kocmi2021nmteval}. ROUGE-L is a widely adopted proxy metric that focuses on the longest common subsequence between the generated text and the reference text, which captures the semantic similarity between the generated and reference texts rather than relying solely on exact word matches. ROUGE-L may not fully capture aspects like fluency, coherence and should be used in conjunction with other metrics and human evaluations to provide a fuller assessment of text generation quality.} for generation tasks. The WikiSQL dataset has its own \href{https://github.com/salesforce/WikiSQL/tree/master}{evaluation suite}, however due to challenges integrating the WikiSQL evaluation suite, we have adopted the ROUGE metric as a proxy for assessing query quality\footnote{Although ROUGE is not tailored for SQL queries, it offers a viable alternative for gauging the alignment between generated and target queries.}. For coding, we use HumanEval~\cite{59chen2021codex}. For GSM8K~\cite{gsm8k}, a regex-based heuristic~\cite{47eval-harness} is used to extract the mathematical answer to be consistent with the Open LLM Leaderboard~\cite{34open-llm-leaderboard}. All metrics are on a 0 to 1 scale, where 0 is the worst possible score, and 1 the best possible score.

Non-fine-tuned models often generate more varied outputs, including unintended artifacts such as additional words or explanations not specified in the prompt. For classification tasks, sometimes these models will generate the actual class string like "Yes/No", "positive/negative" or "True/False" spelled out, instead of the true "1/0" label in the dataset even when instructed. To minimize metric deductions due to response parsing strictness, we first use a regex-based extraction step to map the model’s response to the ground truth vocabulary. If there are multiple matches in the generated text, the first valid match is used. The code for regex-based pre-metric response extractions are available at \href{https://www.github.com/predibase/lora-bakeoff}{github.com/predibase/lora-bakeoff}.

Financial constraints associated with LLM APIs are not trivial. For example, using GPT-4 to assess the complete WikiSQL test set of 15,878 examples would cost approximately \$400, considering the average input (805) and output (16) token counts per example. Such costs can be prohibitive, especially for organizations or researchers operating on limited budgets.

To manage costs while maintaining rigor, we restrict evaluations to the \textbf{first} 1000 examples for datasets with evaluation splits larger than 1000 examples. We acknowledge that this method may introduce selection bias and affect the generalizability of our findings. We recommend that future research considers more expansive evaluations as resources permit.

\section{Results}

\begin{figure}[ht]
    \centering
    \includegraphics[width=\textwidth]{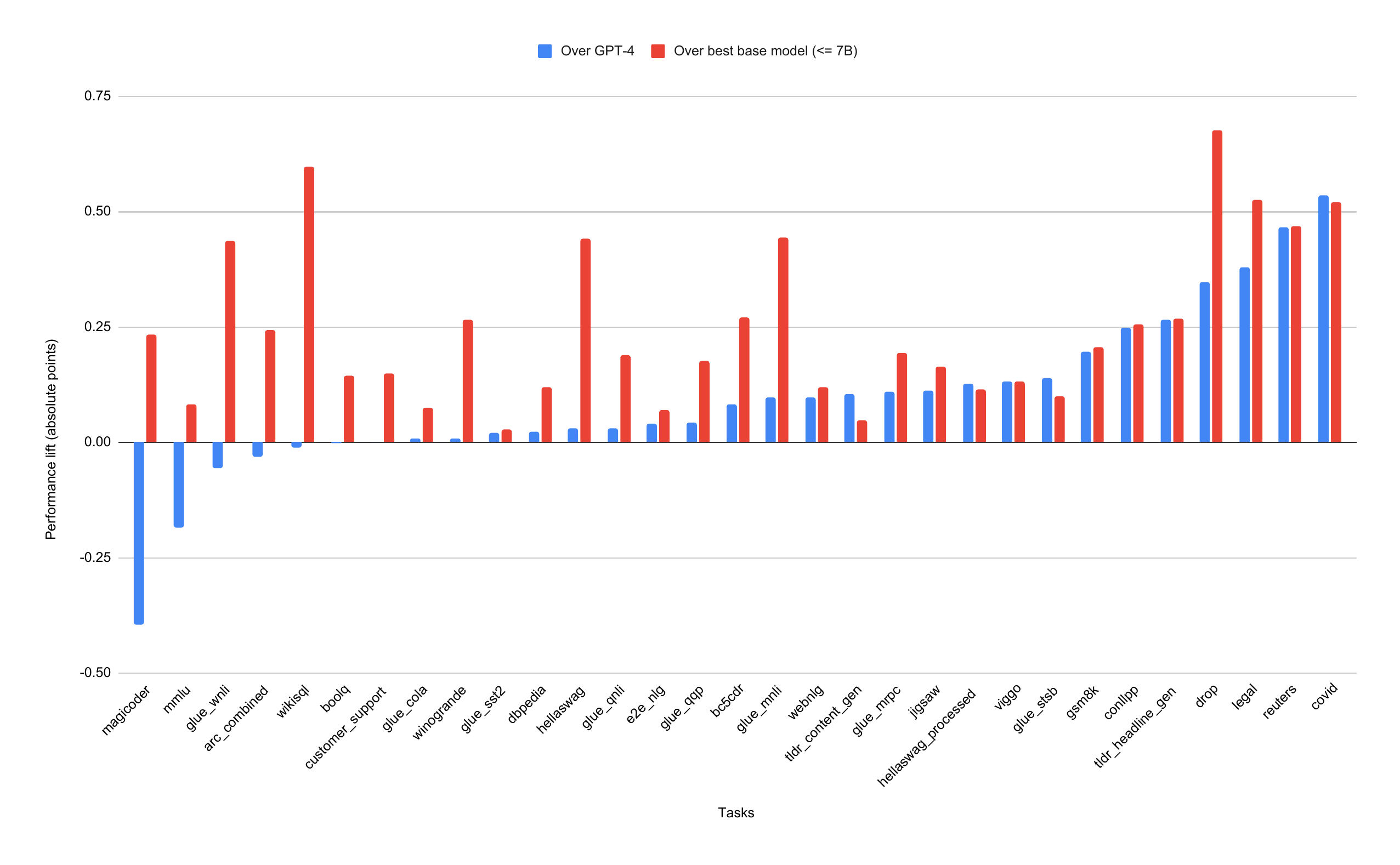}
    \caption{Performance lift from the best fine-tuned LLM over 1) the best base model (<= 7B) (in blue) and GPT-4 (in red) across 31 tasks, in absolute points.}
    \label{fig:finetuninglift}
\end{figure}

LoRA fine-tuning provides a consistent and significant boost from fine-tuning across base models and tasks, as seen in Figure \ref{fig:finetuninglift}. Before fine-tuning, GPT-4 and GPT-3.5 have the strongest performance out of the box compared to all other base models, with 0.599 and 0.661 overall scores, respectively. Performance boosts from fine-tuning range from +26.3 to +51.2 points of improvement depending on the base model, and +38.7 on average (Table \ref{table:performancesummarybytask}). Depending on the task, the best fine-tuned LLM outperforms the best base model from +8.3 to +67.5 points, +25.0 points on average (Table \ref{table:performancesummarybymodel}).

\begin{table*}[ht]
\centering
\footnotesize
\small
\scalebox{0.9}{
\begin{tabular}{ccccccc}
\textbf{Task}        & \textbf{Metric}      & \textbf{Best BM}     & \textbf{Best FT}     & \textbf{GPT-4}       & \textbf{Lift over BM} & \textbf{Lift over GPT-4} \\
\midrule
magicoder            & humaneval            & 0.201                & 0.433                & 0.829                & 0.232                 & -0.396                   \\
mmlu                 & accuracy             & 0.506                & 0.589                & 0.774                & 0.083                 & -0.185                   \\
glue\_wnli           & accuracy             & 0.437                & 0.873                & 0.93                 & 0.436                 & -0.057                   \\
arc\_combined        & accuracy             & 0.673                & 0.915                & 0.947                & 0.242                 & -0.032                   \\
wikisql              & rouge                & 0.301                & 0.898                & 0.909                & 0.597                 & -0.011                   \\
boolq                & accuracy             & 0.764                & 0.909                & 0.911                & 0.145                 & -0.002                   \\
customer\_support    & accuracy             & 0.850                & 1.000                & 1.000                & 0.150                 & 0.000                    \\
glue\_cola           & accuracy             & 0.797                & 0.872                & 0.864                & 0.075                 & 0.008                    \\
winogrande           & accuracy             & 0.576                & 0.84                 & 0.832                & 0.264                 & 0.008                    \\
glue\_sst2           & accuracy             & 0.933                & 0.961                & 0.942                & 0.028                 & 0.019                    \\
dbpedia              & accuracy             & 0.868                & 0.988                & 0.965                & 0.120                 & 0.023                    \\
hellaswag            & accuracy             & 0.393                & 0.834                & 0.805                & 0.441                 & 0.029                    \\
glue\_qnli           & accuracy             & 0.743                & 0.931                & 0.902                & 0.188                 & 0.029                    \\
e2e\_nlg             & rouge                & 0.482                & 0.552                & 0.513                & 0.070                 & 0.039                    \\
glue\_qqp            & accuracy             & 0.708                & 0.883                & 0.841                & 0.175                 & 0.042                    \\
bc5cdr               & rouge                & 0.703                & 0.972                & 0.89                 & 0.269                 & 0.082                    \\
glue\_mnli           & accuracy             & 0.455                & 0.899                & 0.803                & 0.444                 & 0.096                    \\
webnlg               & rouge                & 0.563                & 0.681                & 0.583                & 0.118                 & 0.098                    \\
tldr\_content\_gen   & rouge                & 0.183                & 0.23                 & 0.125                & 0.047                 & 0.105                    \\
glue\_mrpc           & accuracy             & 0.694                & 0.887                & 0.777                & 0.193                 & 0.11                     \\
jigsaw               & accuracy             & 0.704                & 0.867                & 0.754                & 0.163                 & 0.113                    \\
hellaswag\_processed & rouge                & 0.146                & 0.261                & 0.134                & 0.115                 & 0.127                    \\
viggo                & rouge                & 0.374                & 0.505                & 0.374                & 0.131                 & 0.131                    \\
glue\_stsb           & mae                  & 0.814                & 0.913                & 0.773                & 0.099                 & 0.14                     \\
gsm8k                & accuracy             & 0.364                & 0.569                & 0.373                & 0.205                 & 0.196                    \\
conllpp              & rouge                & 0.733                & 0.989                & 0.742                & 0.256                 & 0.247                    \\
tldr\_headline\_gen  & rouge                & 0.174                & 0.441                & 0.175                & 0.267                 & 0.266                    \\
drop                 & rouge                & 0.066                & 0.741                & 0.393                & 0.675                 & 0.348                    \\
legal                & rouge                & 0.158                & 0.683                & 0.305                & 0.525                 & 0.378                    \\
reuters              & rouge                & 0.010                & 0.479                & 0.014                & 0.469                 & 0.465                    \\
covid                & accuracy             & 0.322                & 0.843                & 0.309                & 0.521                 & 0.534                    \\
\multicolumn{1}{l}{} & \multicolumn{1}{l}{} & \multicolumn{1}{l}{} & \multicolumn{1}{l}{} & \multicolumn{1}{l}{} & \multicolumn{1}{l}{}  & \multicolumn{1}{l}{}     \\
\midrule
Average              & \multicolumn{1}{l}{} & \textbf{0.506}       & \textbf{0.756}       & \textbf{0.661}       & \textbf{0.250}        & \textbf{0.095}          
\end{tabular}
}

\caption{Best model performance for each task, before and after fine-tuning, compared to GPT-4.}
\label{table:performancesummarybytask}

\end{table*}

After fine-tuning, 301/310 models surpass their base model counterpart\footnote{Most instances where fine-tuning was worse than the base model were in the family of Gemma models. This is possibly due to the bugs with the Gemma family of models as identified by Unsloth\cite{65unslothUnslothFixing}, which were not accounted for when benchmarks were collected.}, while 224/310 fine-tuned LLMs surpass the benchmark set by GPT-4 (Table \ref{table:performancesummarybymodel}). Gemma-2b is the worst performing base model after fine-tuning, but also experiences the largest lift from fine-tuning overall, which suggests that models with lower initial scores stand to benefit the most from fine-tuning (Figure 1).

By overall average across all tasks, all fine-tuned models perform better than GPT-3.5, and all 7B fine-tuned models perform better than GPT-4, except for gemma-7b and gemma-7b-it. Phi-2, with as few as 2 billion parameters, exhibits performance competitive with GPT-4 after fine-tuning, consistent with the findings of the Phi-2 technical report [46].

Averaged over 31 tasks, the overall performance of the best fine-tuned LLMs (0.756) are significantly higher than GPT-4 (0.661) (Table \ref{table:performancesummarybymodel}). A detailed breakdown of performance per model, per task, can be found in Appendix \ref{sec:appendixresults}.

\begin{table*}[ht]
\centering
\footnotesize
\small
\scalebox{0.70}{
\begin{tabular}{cccccccc}
\textbf{Base Model}  & \textbf{No FT}       & \textbf{With FT}     & \textbf{\begin{tabular}[c]{@{}c@{}}Average lift\\ from FT\end{tabular}} & \textbf{\begin{tabular}[c]{@{}c@{}}Average lift\\ from FT\\ vs. GPT-4\end{tabular}} & \textbf{\begin{tabular}[c]{@{}c@{}}Frequency \\ FT \textgreater No FT\end{tabular}} & \textbf{\begin{tabular}[c]{@{}c@{}}Frequency\\ FT \textgreater GPT-4\end{tabular}} & \textbf{\begin{tabular}[c]{@{}c@{}}Frequency\\ FT = max(task)\end{tabular}} \\
\midrule
gpt-3.5-turbo        & 0.599                & ---                  & ---                                                                     & ---                                                                                 & ---                                                                                 & ---                                                                                & 0/31                                                                        \\
gemma-2b-instruct    & 0.326                & 0.645                & 0.319                                                                   & -0.016                                                                              & 96.7\% (30/31)                                                                      & 64.5\% (20/31)                                                                     & 0/31                                                                        \\
gemma-7b             & 0.187                & 0.645                & 0.458                                                                   & -0.016                                                                              & 93.5\% (29/31)                                                                      & 64.5\% (20/31)                                                                     & 1/31                                                                        \\
gemma-7b-instruct    & 0.377                & 0.656                & 0.279                                                                   & -0.005                                                                              & 83.8\% (26/31)                                                                      & 64.5\% (20/31)                                                                     & 0/31                                                                        \\
gemma-2b             & 0.145                & 0.657                & \textbf{0.512}                                                          & -0.004                                                                              & 100.0\% (31/31)                                                                     & 67.7\% (21/31)                                                                     & 0/31                                                                        \\
gpt-4                & \textbf{0.661}       & ---                  & ---                                                                     & ---                                                                                 & ---                                                                                 & ---                                                                                & 6/31                                                                        \\
phi-2                & 0.274                & 0.677                & 0.403                                                                   & 0.016                                                                               & 100.0\% (31/31)                                                                     & 71.0\% (22/31)                                                                     & 1/31                                                                        \\
llama-2-7b           & 0.252                & 0.696                & 0.444                                                                   & 0.035                                                                               & 96.7\% (30/31)                                                                      & 67.7\% (21/31)                                                                     & 0/31                                                                        \\
llama-2-7b-chat      & 0.370                & 0.708                & 0.337                                                                   & 0.047                                                                               & 100.0\% (31/31)                                                                     & 74.2\% (23/31)                                                                     & 0/31                                                                        \\
mistral-7b-instruct  & 0.462                & 0.724                & 0.263                                                                   & 0.063                                                                               & 100.0\% (31/31)                                                                     & 77.4\% (24/31)                                                                     & 3/31                                                                        \\
mistral-7b           & 0.271                & 0.732                & 0.461                                                                   & 0.071                                                                               & 100.0\% (31/31)                                                                     & 83.8\% (26/31)                                                                     & \textbf{10/31}                                                              \\
zephyr-7b-beta       & 0.350                & \textbf{0.742}       & 0.392                                                                   & 0.081                                                                               & 100.0\% (31/31)                                                                     & 87.1\% (27/31)                                                                     & 8/31                                                                        \\
\multicolumn{1}{l}{} & \multicolumn{1}{l}{} & \multicolumn{1}{l}{} & \multicolumn{1}{l}{}                                                    & \multicolumn{1}{l}{}                                                                & \multicolumn{1}{l}{}                                                                & \multicolumn{1}{l}{}                                                               & \multicolumn{1}{l}{}                                                        \\
\midrule
Average              & 0.301                & 0.688                & 0.387                                                                   & 0.027                                                                               & 97.1\% (301/310)                                                                    & 72.3\% (224/310)                                                                   & \multicolumn{1}{l}{}                                                       
\end{tabular}
}
\caption{Model performance by base model averaged over 31 tasks, before and after fine-tuning.}
\label{table:performancesummarybymodel}
\end{table*}


\section{Discussion and Analysis}

\subsection{Which Base Model is the best for LoRA Fine-tuning?}

\begin{figure}[ht]
    \centering
    \scalebox{0.7}{
    \includegraphics[width=\textwidth]{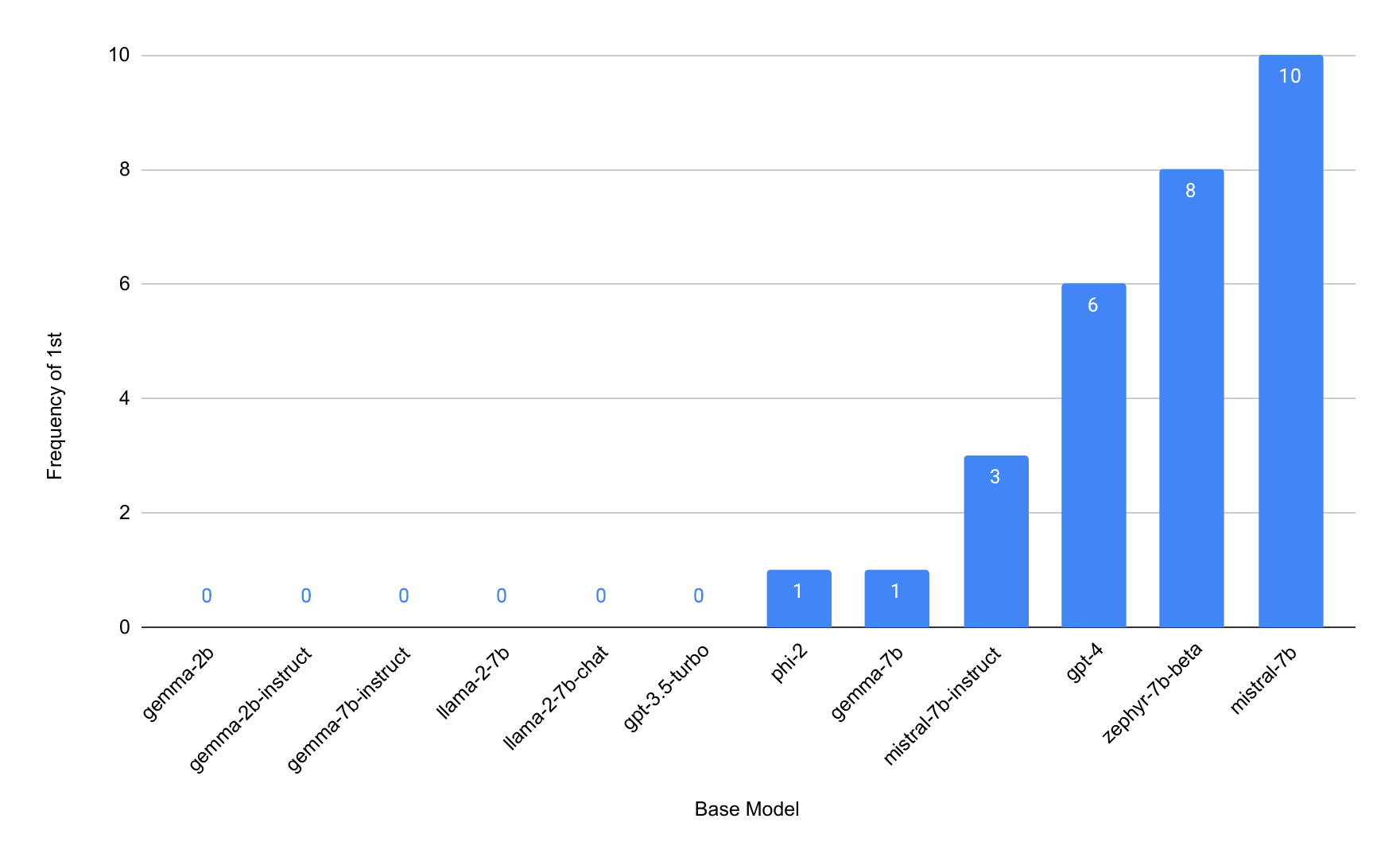}
    }
    \caption{Frequency of base models (with fine-tuning) as the top performer for a task. Ties, namely for the \textit{customer\_support} task where most models attain 100\% perfect scores, are excluded.}
    \label{fig:topperformer}
\end{figure}

Mistral-7B and Zephyr-7b-beta emerge as leaders, albeit in different categories. Mistral-7B frequently achieves top performance across the most number of tasks (10/31), suggesting a high adaptability (Figure \ref{fig:topperformer}). Conversely, Zephyr boasts the highest overall average performance (0.731). Mistral-7b, Mistral-7b-instruct, and Zephyr-7b-beta (which is itself based on Mistral-7b-instruct~\cite{66zephyr}) lead the pack for LoRA fine-tuning performance, ahead of Llama, Phi, and Gemma families.

\subsection{Does size matter for LoRA fine-tuning? 2B vs. 7B}

The 2B parameter Phi-2 model, after fine-tuning, outperforms all of the 2B and 7B Gemma models by overall overage, and is only 1.9 points behind the next highest performing 7B model, Llama-2-7b (0.677 vs. 0.696). Despite this, we find that fine-tuned 7B models are almost always better than fine-tuned 2B models (29/31 tasks). Among 2B parameter models in particular (Phi and Gemma), we see that all Gemma instruct models were better than Phi out of the box, however, Phi-2 performs better than all other Gemma models after fine-tuning.

\subsection{Is fine-tuning better with Instruction-tuned or Auto-complete models?}

\begin{figure}[ht]
    \centering
    \scalebox{0.7}{
    \includegraphics[width=\textwidth]{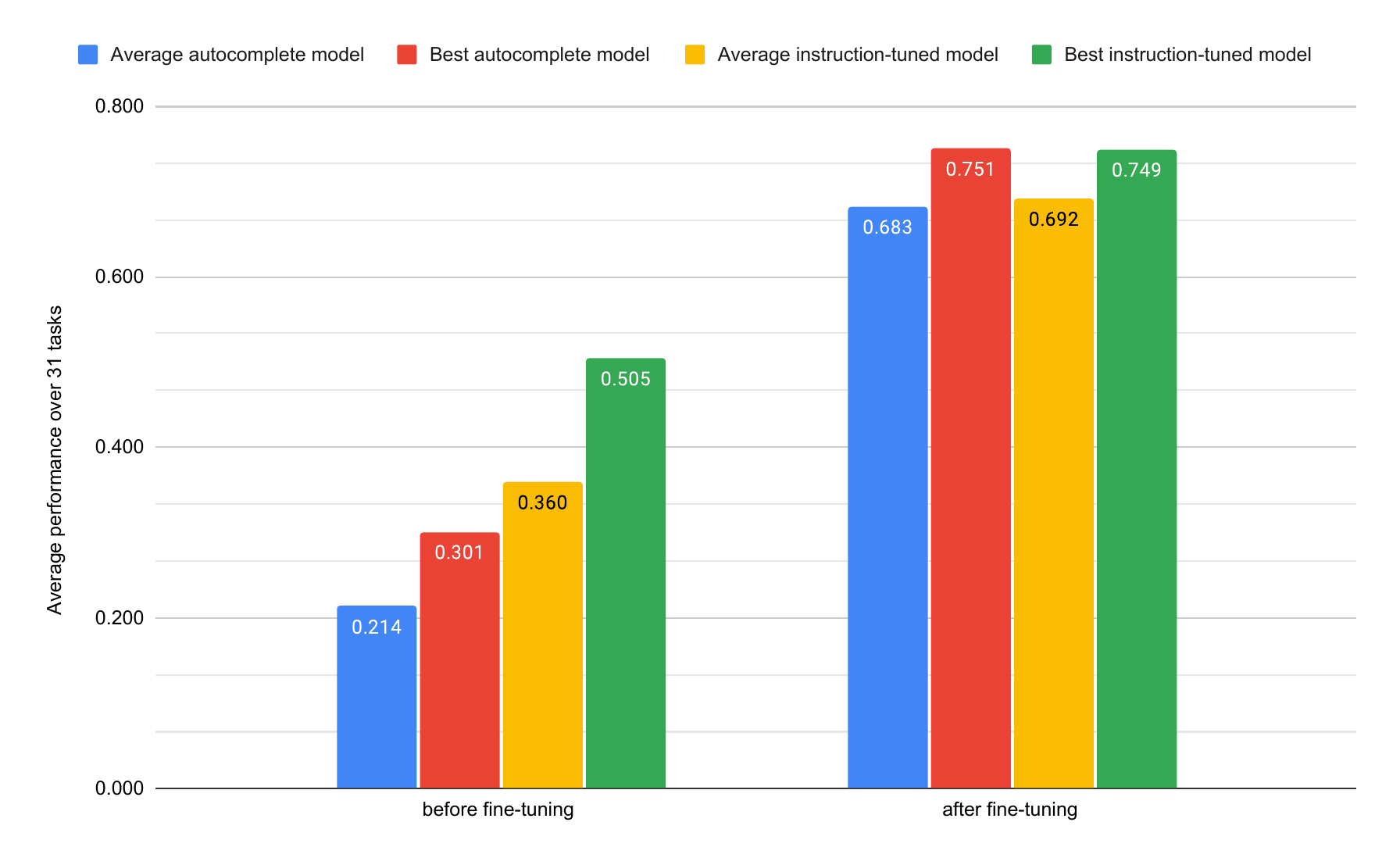}
    }
    \caption{Comparison of auto-complete vs. instruction-tuned base models, before and after fine-tuning.}
    \label{fig:autocompletevsinstructiontuned}
\end{figure}

In Figure \ref{fig:autocompletevsinstructiontuned}, we observe that before fine-tuning, instruction-tuned models outperform auto-complete models, despite using completion style prompts. A qualitative analysis shows that auto-complete models were much more likely to "go off the rails", and generate long irrelevant text sequences, and instruction-tuned models demonstrate a higher consistency in correctly attempting the imminent task.

After fine-tuning, the performance disparities between the models narrow. The average instruction-tuned model slightly outperforms the average auto-complete model by a margin of +0.009, however the reverse is true when comparing the best fine-tuned instruction-tuned model and the best fine-tuned auto-complete model (-0.002). Auto-complete models, possibly due to their broader and less specialized knowledge base, may be inherently more adaptable to a variety of tasks. However, with adequate fine-tuning, both types of models achieve comparable performance levels. We encourage further research to explore how the foundational design of instruction-tuned models influences their adaptability and effectiveness in task-specific fine-tuning.

\subsection{When does GPT-4 consistently outperform fine-tuned models?}

We observe a distinct advantage for fine-tuned LLMs on narrowly-scoped tasks, such as those within the GLUE benchmarks. These tasks, primarily classification-oriented, saw fine-tuned LLMs achieve near 90\% accuracy, outperforming GPT-4. GPT-4 continues to outperform fine-tuned models in 6 out of 31 tasks, particularly in broader, more complex domains such as Python coding and MMLU.

\subsection{Quantifying the relationship between fine-tuning quality lift and task complexity}

If fine-tuned models perform better on specialized "narrow" tasks and worse on "broader" tasks,  can we establish a predictive relationship between the complexity of a task and the efficacy of LoRA fine-tuning?  Identifying such a relationship could provide a valuable predictive tool for assessing the potential benefits of fine-tuning enhancements on new tasks before the fine-tuning process begins.

\subsubsection{Heuristics for fine-tuning quality, quality lift, and task complexity}

To quantify task complexity, we use several heuristics:

\begin{itemize}
    \item Number of training examples
    \item Lengths of inputs and outputs ($\mu$, $\sigma$, and 95th percentile).
    \item Compressibility\footnote{https://docs.python.org/3/library/gzip.html} ($\mu$ and $\sigma$)
    \item Diversity of content, which we approximate by measuring the rouge-L similarity between inputs and outputs)~\cite{50wang2022selfinstruct} ($\mu$ and $\sigma$).
\end{itemize}

For task complexity heuristic, 

For model quality measurements, we track:

\begin{itemize}
    \item Baseline GPT-4 score
    \item Lift from the best fine-tuned model vs. GPT-4 ("Max GPT-4 Lift")
    \item Average fine-tuning lift over the base model
    \item Best base model score without fine-tuning
    \item Average base model score without fine-tuning
    \item Best fine-tuned model score
    \item Average fine-tuned model score

\end{itemize}

Refer to Table \ref{table:datasetcomplexityheuristics} for a complete example.

\begin{table*}[ht]

\scalebox{0.9}{
\begin{tabular}{clccc}
\multicolumn{1}{l}{}                            & \textbf{Metric}          & \multicolumn{1}{l}{\textbf{arc\_combined}} & \multicolumn{1}{l}{\textbf{bc5cdr}} & \multicolumn{1}{l}{\textbf{boolq}} \\
\midrule
\multirow{6}{*}{Model quality measurements}     & Max GPT-4 Lift           & -0.03                                      & 0.08                                & 0.00                               \\
                                                & Average Base Model Lift  & 0.32                                       & 0.75                                & 0.19                               \\
                                                & Best Base Model Score    & 0.67                                       & 0.70                                & 0.76                               \\
                                                & Average Base Model Score & 0.41                                       & 0.22                                & 0.64                               \\
                                                & Best Fine-tuned Score    & 0.92                                       & 0.97                                & 0.91                               \\
                                                & Average Fine-Tuned Score & 0.73                                       & 0.97                                & 0.82                               \\
\midrule
\multirow{14}{*}{Task complexity heuristics} & Input length p95         & 143.00                                     & 175.00                              & 270.70                             \\
                                                & Input length $\mu$           & 102.89                                     & 142.15                              & 145.23                             \\
                                                & Input length $\sigma$           & 21.68                                      & 19.17                               & 69.03                              \\
                                                & Output length p95        & 1.00                                       & 58.00                               & 1.00                               \\
                                                & Output length $\mu$          & 1.00                                       & 37.11                               & 1.00                               \\
                                                & Output length $\sigma$          & 0.00                                       & 11.27                               & 0.00                               \\
                                                & Example length $\mu$         & 102.92                                     & 178.26                              & 146.23                             \\
                                                & Example length p95       & 143.00                                     & 226.05                              & 271.70                             \\
                                                & Example length $\sigma$         & 21.66                                      & 27.84                               & 69.03                              \\
                                                & I/O rougeL similarity $\mu$  & 0.03                                       & 0.19                                & 0.00                               \\
                                                & I/O rougeL similarity $\sigma$  & 0.01                                       & 0.03                                & 0.00                               \\
                                                & Compressibility $\mu$        & 0.64                                       & 0.55                                & 0.60                               \\
                                                & Compressibility $\sigma$        & 0.06                                       & 0.01                                & 0.07                               \\
                                                & \# training examples     & 3370                                       & 5228                                & 9427                              
\end{tabular}
}
\caption{Model quality measurements and task complexity heuristics for 3 different tasks (example). Refer to the Appendix \protect\ref{sec:appendixresults}.
\label{table:datasetcomplexityheuristics} for all measurements and heuristics for all 31 tasks.}

\end{table*}

\subsubsection{Correlating fine-tuning quality and quality lift with task complexity}

We find several intriguing correlations suggesting significant interactions between our task complexity heuristics and measurements of model performance. Key observations include:

\begin{itemize}

\item \textbf{Compressibility} exhibited a dual influence, correlating positively with both best and average base model scores (0.36), while correlating negatively with these scores when the variance in compressibility increased (-0.37). This indicates that while uniform compressibility supports model performance, higher variability in compressibility tends to degrade it.
\item \textbf{Input and Output Lengths:} Longer and more varied output lengths correlated positively with the maximum lift from GPT-4 fine-tuning, suggesting that tasks with extended and more varied outputs are not detrimental for fine-tuning. Conversely, longer and more varied input and output lengths negatively correlate with absolute base and fine-tuned model scores.
\item \textbf{Input and Output Rouge-L Similarity:} A higher standard deviation in input/output Rouge-L similarity correlates negatively with both base and fine-tuned model scores. This suggests that greater variability in content similarity within a dataset may pose difficulties for model learning.
\item \textbf{Number of training examples:} No significant correlation was found with the number of training examples, pointing to the possibility that once a sufficient sample size is achieved, additional examples do not necessarily contribute to improved fine-tuning efficacy.
\item \textbf{Model quality inter-correlations} reveal that better average scores (both base and fine-tuned) strongly predict the best scores obtained, suggesting a general consistency in model performance across different training instances.

\end{itemize}

Overall, these observations are consistent with our hypothesis that narrower easier tasks are more likely to see success with fine-tuned adapters.

\begin{figure}[ht]
    \label{fig:taskcorrelations}
    \centering
    \includegraphics[width=\textwidth]{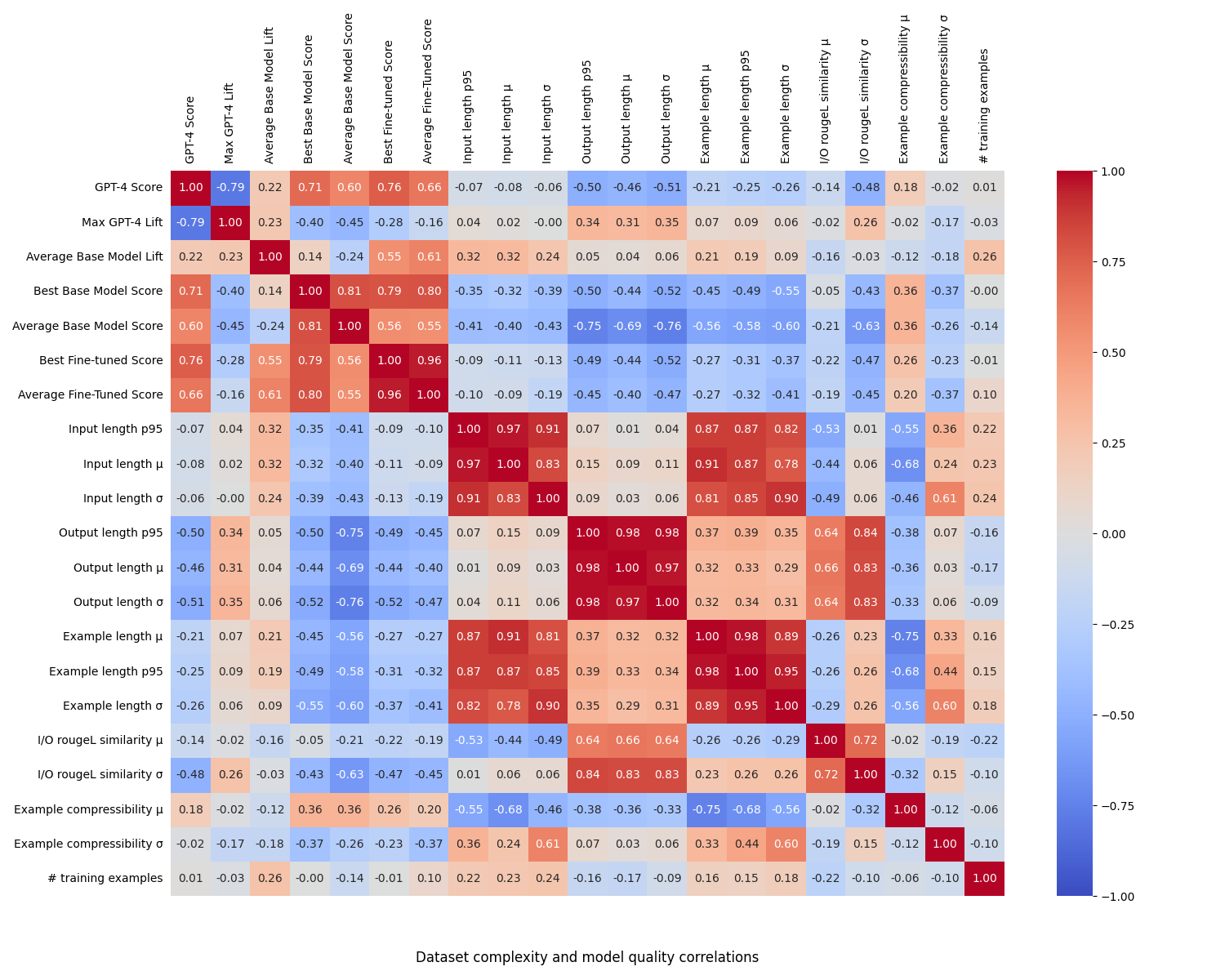}
    \caption{Correlations between dataset complexity and model quality correlations for 310 LLMs across 31 tasks, before and after LoRA-based fine-tuning.}
\end{figure}

\subsubsection{Predicting fine-tuning quality and quality lift given task complexity heuristics}

We train linear regression models to predict the quality lift achievable through adapter-based fine-tuning, using z-score normalized dataset complexity heuristics (described in Table \ref{table:datasetcomplexityheuristics}) as predictors. Results are summarized in Table \ref{table:liftpredictionresults}, where we find that linear models yield root mean squared errors (RMSE) of 0.166 to 0.092, depending on the model quality metric in question.

Incorporating the score of the average base model without fine tuning as an additional feature improves prediction accuracy for all model quality metrics (+0.004 to +0.069). This demonstrates some predictive power in knowing base model performance for anticipating potential gains from fine-tuning. RMSE errors are rather low, suggesting that upfront heuristics-based measurements of dataset complexity can be reasonable indicators of positive fine-tuning impact.

\begin{table*}[ht]
\centering
\scalebox{0.8}{
\begin{tabular}{ccc}
\textbf{Model Quality Metric} & \textbf{\begin{tabular}[c]{@{}c@{}}With average base model score \\ as a feature \\ (RMSE)\end{tabular}} & \textbf{\begin{tabular}[c]{@{}c@{}}With average base model score\\ as a feature\\ (RMSE)\end{tabular}} \\
\midrule
GPT-4 Score                   & 0.140                                                                                                    & 0.121                                                                                                  \\
Max GPT-4 Lift                & 0.092                                                                                                    & 0.085                                                                                                  \\
Average Base Model Score      & 0.099                                                                                                    & N/A (0.000)                                                                                            \\
Best Base Model Score         & 0.166                                                                                                    & 0.097                                                                                                  \\
Average Base Model Lift       & 0.099                                                                                                    & 0.095                                                                                                  \\
Average Fine-Tuned Score      & 0.119                                                                                                    & 0.095                                                                                                  \\
Best Fine-tuned Score         & 0.097                                                                                                    & 0.091                                                                                                 
\end{tabular}
}

\caption{The performance of linear regression models predicting model quality heuristics before and after fine-tuning, given z-score normalized dataset complexity heuristics, with and without a representative base model score.}
\label{table:liftpredictionresults}

\end{table*}
\section{Performance Benchmarks of LoRAX Deployments}

To assess the viability of serving many LoRA fine-tuned LLMs simultaneously in a real-world application, we launch LoRA Land. LoRA Land is a web application that serves 25 fine-tuned Mistral-7b LLMs served to thousands of users from a single A100 GPU.

\begin{figure}[ht]
    \centering
    \includegraphics[width=\textwidth]{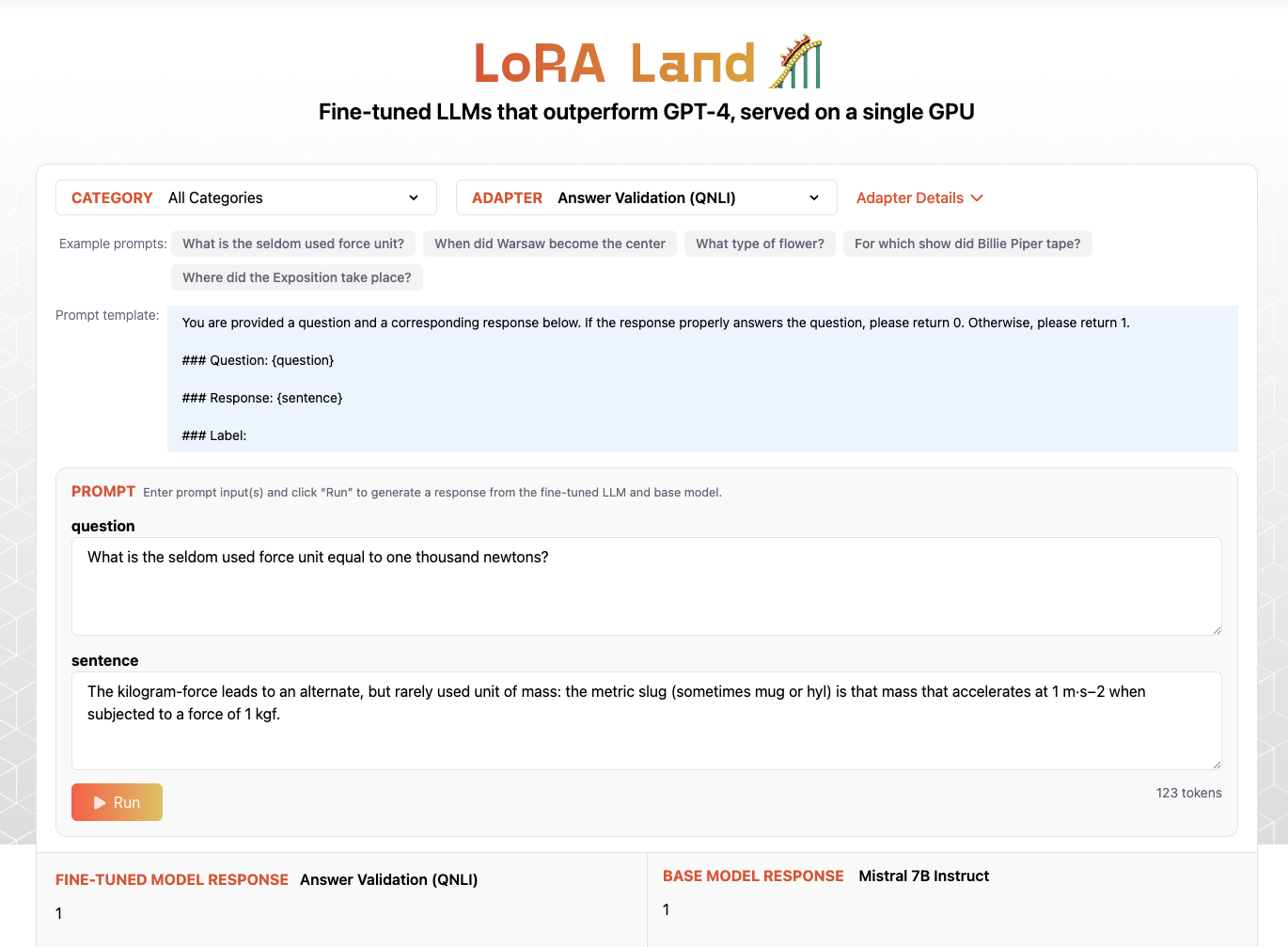}
    \vspace{5mm}
    \caption{The LoRA Land web application that serves 25 fine-tuned LLMs on a single A100. The application is available at \href{https://predibase.com/lora-land}{https://predibase.com/lora-land}.}
    \label{fig:loraland}
\end{figure}

\subsection{LoRAX in a Nutshell}

LoRA Exchange (LoRAX) \cite{12predibaseLoRAExchange} is an open source Multi-LoRA inference server specifically designed for serving many fine-tuned models at once using a shared set of GPU resources. Compared with conventional dedicated LLM deployments, LoRAX consists of three novel components:

\begin{itemize}
    \item \textbf{Dynamic Adapter Loading}, allowing each set of fine-tuned LoRA weights to be loaded from storage just-in-time as requests come in at runtime, without blocking concurrent requests.

    \item \textbf{Continuous Multi-Adapter Batching}, a fair scheduling policy for optimizing aggregate throughput of the system that extends the popular continuous batching strategy to work across multiple sets of LoRA adapters in parallel.

    \item \textbf{Tiered Weight Caching}, to support fast exchanging of LoRA adapters between requests, and offloading of adapter weights to CPU and disk to avoid out-of-memory errors.

\end{itemize}

\begin{figure}[ht]
    \centering
    \includegraphics[width=\textwidth]{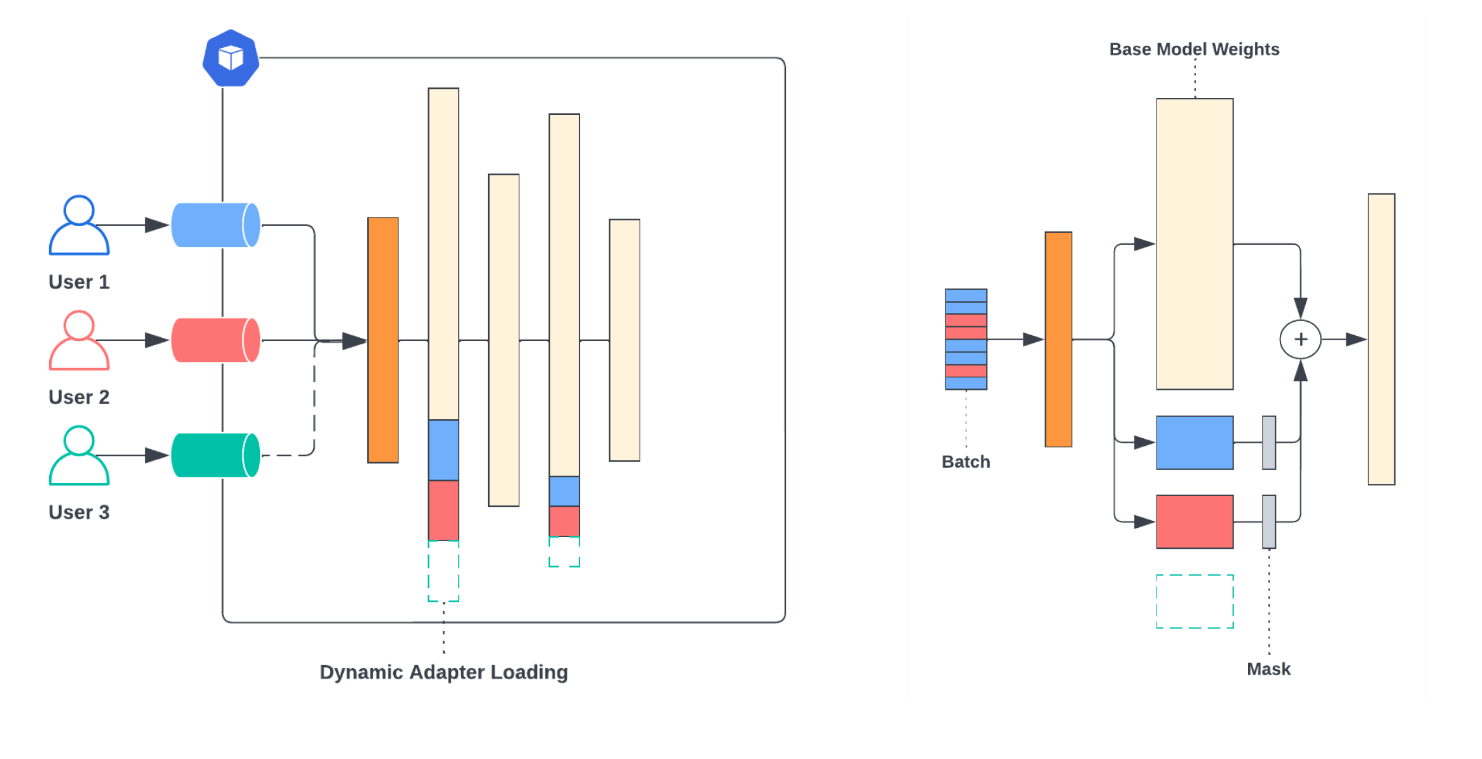}
    \caption{\textbf{Dynamic adapter loading} (left) enables multiple concurrent fine-tuned models to process requests. User 3’s model (green) is loaded in the background while the other requests proceed as usual. \textbf{Continuous Multi-Adapter Batching} (right): Multiple adapters are decoded in a single batch. Masks ensure that only the right adapter is used for processing each element of the batch.}
    \label{fig:loraxnutshell}
\end{figure}

\subsection{Benchmarking Results}

We run benchmarks in order to understand the impact of serving multiple adapters on the relevant metrics, described below. We also test the scalability of the system with respect to the following factors:

\begin{itemize}
\item Number of concurrent users submitting LLM prompts
\item Number of adapters concurrently being queried
\item Number of input tokens
\item Number of output tokens
\end{itemize}

LLM serving performance metrics include: time to first token (TFTT), total request time, token streaming time, and throughput (tokens per second). We run our benchmarks from a t3.2xlarge EC2 instance in the AWS zone us-west-2. All benchmarks are based on the Mistral-7b-instruct LLM, deployed on an A100 GPU with 80GB of RAM. The script used to benchmark LLM serving performance can be found in Appendix \ref{sec:appendixscripts}.

The following is a summary of relevant terminology:

\begin{itemize}

\item \textbf{Total request time (ms):} total time from when the request is sent to when the last token is streamed back to the client.
\item \textbf{Time to first token, TTFT (ms):} time from when the request is sent to the first token is received by the client
\item \textbf{Token streaming time (ms):} time from when the first token is received by the client to when the last token is received by the client.
\item \textbf{Throughput (token/s):} number of tokens generated per seconds, computed by (Token streaming time (ms) / number of output tokens)
\item \textbf{Concurrent users:} number of users that make requests to the LLM, wait until they receive a full response, then make another request until the end of testing time.

\end{itemize}

\subsection{Latency from adapter switching and concurrent users}

The following reported benchmarks come from 2-minute runs that continuously stream requests to the LLM deployment. Our experiments indicate that a duration of two minutes provides an adequate volume of data to obtain stable and reliable metrics.

Table \ref{table:adapterswitching} shows the impact LLM query performance isolated to adapter switching mechanics. In the multi-adapter, multi-user case, we see that the token streaming time is the same, but the total request time differs by 7.21ms which illustrates the cost of handling requests from 100 concurrent users that lead to switching between 25 adapters.

\begin{table*}[ht]
\centering

\scalebox{0.7}{
\begin{tabular}{rcccc}
\multicolumn{1}{c}{}           & \multicolumn{2}{c}{0 adapters (base model), 1 concurrent user} & \multicolumn{2}{c}{25 adapters (base model), 100 concurrent user} \\
\cmidrule(l{3pt}r{3pt}){2-3}\cmidrule(l{3pt}r{3pt}){4-5}
\multicolumn{1}{c}{}           & Average                        & p90                           & Average                          & p90                            \\
\midrule
Total request time (ms)        & 191.81                         & 192.3                         & 199.02                           & 201.82                         \\
Time to first token, TTFT (ms) & 122.19                         & 191.16                        & 128.79                           & 199.11                         \\
Token streaming time (ms)      & 70                             & 92.38                         & 70.14                            & 96.62                         
\end{tabular}
}

\caption{Measuring LLM querying metrics from adapter switching mechanics only. To eliminate extra, non-adapter-switching factors related to input and generation, simulated requests contain 1 input token and \textit{max\_new\_tokens} is capped at 1. Throughput metrics are excluded since only 1 output token is generated. \label{table:adapterswitching}}
\end{table*}

\begin{table}[ht]
\centering  
\footnotesize 

\begin{tabular}{ccccccc}
\# concurrent users                             &         & 1       & 5       & 10      & 20      & 50      \\
\midrule
\multirow{2}{*}{Total request time (ms)}        & average & 943.03  & 1165.71 & 1359.39 & 2004.9  & 2981.66 \\
                                                & p90     & 1567.66 & 1925.96 & 2147.84 & 3287.21 & 4673.52 \\
\multirow{2}{*}{Time to first token, TTFT (ms)} & average & 121.84  & 121.80  & 143.68  & 135.43  & 136.17  \\
                                                & p90     & 191.08  & 195.85  & 199.98  & 199.76  & 199.54  \\
\multirow{2}{*}{Token streaming time (ms)}      & average & 821.09  & 1043.79 & 1215.6  & 1869.36 & 2845.38 \\
                                                & p90     & 1468.76 & 1804.16 & 2007.89 & 3130.72 & 4544.64
\end{tabular}
\vspace{5mm}
\caption{Benchmarking base LLM deployments on 1xA100 with queries that simulate real load.}
\end{table}

To simulate realistic traffic payloads, we generate random payloads with 30-500 input tokens and 1-120 output tokens, modeled off of the tasks defined in Table \ref{table:datasets}. We vary the number of concurrent users from 1 to 50, and payloads are issued randomly between 25 different adapter endpoints. 

When scaling from 1 to 50 concurrent users, which also increases load by 50X, the average time to first token (TTFT) is slightly affected (+21.84ms or 17.9\% increase). We see a 3.46X decrease in throughput for the same 50X increase in load.

\begin{table}[ht]

\scalebox{0.9}{
\begin{tabular}{lcccccc}
\# concurrent users                             & \multicolumn{1}{l}{} & 1       & 5       & 10      & 20      & 50      \\
\midrule
\multirow{2}{*}{Total request time (ms)}        & average              & 956.56  & 1272.16 & 1528.99 & 1896.1  & 3336.27 \\
                                                & p90                  & 1758.53 & 2164.08 & 2612.05 & 3222.73 & 5330.84 \\
\multirow{2}{*}{Time to first token, TTFT (ms)} & average              & 170.62  & 148.14  & 157.49  & 167.28  & 153.89  \\
                                                & p90                  & 199.36  & 198.98  & 199.41  & 200.99  & 200.2   \\
\multirow{2}{*}{Token streaming time (ms)}      & average              & 785.82  & 1123.91 & 1371.39 & 1728.71 & 3182.27 \\
                                                & p90                  & 1594.65 & 2023.33 & 2468.87 & 3047.92 & 5169.05
\end{tabular}
}

\vspace{5mm}
\caption{Benchmarking 25 adapters on 1xA100 with queries that simulate real load. \label{table:loraxconcurrentusers}}

\end{table}

Table \ref{table:loraxconcurrentusers} shows that there’s no significant difference between querying the base LLM vs. the 25 adapters when it comes to TTFT or throughput. The cost of adapter switching is overshadowed by the time it takes to generate tokens once requests come in. Comparing average case numbers vs. p90 numbers for TTFT, the largest disparity is between 121.8ms (average) and 195.95ms (p90) for a 60.87\% increase. Additionally, we consistently see that TTFT is at or under the 200ms mark.

On throughput, we observe that it takes between 12 and 13.5ms to generate a single token on an A100 GPU both for base deployments and deployments where adapter weights have been added. This means that the aggregate throughput for the LLM deployment on that GPU is between 74 tokens/s and 83 tokens/s.

\subsection{Analyzing the performance impact of additional deployment replicas}

In Table \ref{table:loraxreplicas}, we run benchmarks for 25 adapters queried concurrently by 50 users, with a LoRAX deployment on 1 replica. We then run benchmarks where we scale the LoRAX deployment to 2 replicas placed behind a round robin load balancer to route equal amounts of traffic to each replica, while also scaling the load to 100 concurrent users. We see that the numbers are stable across the board, signifying that replicas can be scaled linearly with load to achieve comparable metrics.

\begin{table}[ht]
\scalebox{0.8}{
\begin{tabular}{lccc}
                                                & \multicolumn{1}{l}{} & 50 Concurrent users, 1 replica & 100 Concurrent users, 2 replicas \\
\midrule
\multirow{2}{*}{Total request time (ms)}        & average              & 3336.27                        & 3368.53                          \\
                                                & p90                  & 5330.84                        & 5382.61                          \\
\multirow{2}{*}{Time to first token, TTFT (ms)} & average              & 153.89                         & 161.97                           \\
                                                & p90                  & 200.2                          & 199.83                           \\
\multirow{2}{*}{Token streaming time (ms)}      & average              & 3182.27                        & 3206.46                          \\
                                                & p90                  & 5169.05                        & 5248.97                         
\end{tabular}
}

\vspace{5mm}
\caption{Benchmarking 25 adapters on 1 LoRAX replica vs. 2 replicas with queries that simulate real load.}
\label{table:loraxreplicas}
\end{table}
\section{Limitations}
\label{sec:limitations}

Our experimental design has many limitations, including:

\begin{itemize}

\item \textbf{Restricted Evaluation Scope:} Our evaluations are limited to the first 1000 examples of datasets with larger evaluation splits to manage costs while maintaining rigor. This may introduce selection bias and limit the generalizability of our findings. Future research should consider more comprehensive evaluations as resources allow.

\item \textbf{Prompt Engineering Constraints:} Our study does not employ advanced prompt engineering techniques such as majority voting, n-shot prompting, or specialized tuning methods like MedPrompt or chain-of-thought prompting. In this study, we prioritize reproducibility and minimize biases from selective example choice by using simple zero or single-shot prompts across all tasks, however these techniques have shown potential in enhancing task-specific performance.

\item \textbf{Training Constraints:} All LLMs are fine-tuned with the same Models are trained with consistent parameters: 40K examples, batch size of 1, 4-bit quantization, and a LoRA rank of 8, using an adam optimizer and a cosine learning rate scheduler with specific settings. Training is conducted on a single A10 GPU, using gradient checkpointing to manage memory limitations. For datasets where full sequence lengths induce memory overflow, we truncate sequences to the 95th percentile length. This approach may impact the thoroughness of model training, particularly on datasets where 40K steps do not complete even one full epoch. Expanding hardware capabilities, increasing batch sizes, or adjusting hyperparameters like the learning rate or scheduler could potentially enhance outcomes.

\item \textbf{Limited Model Variety:} Our experiments are limited to LoRA fine-tuning on two model sizes, 2B and 7B. Exploring a broader range of model sizes, including larger models such as 13B or 70B, could provide insights into the scalability and effectiveness of fine-tuning across different computational capacities.

\end{itemize}

We maintain that LoRA Land successfully demonstrates the practical efficiency of training and serving several task-specialized LLMs that rival GPT-4 in a production application powered by LoRAX, despite these limitations.

\section{Conclusion}
In this study, we assess the efficacy of Low Rank Adaptation (LoRA) for fine-tuning Large Language Models (LLMs) across a broad range of tasks and models and the viability of serving multiple fine-tuned LoRA LLMs in production.

On model quality, our results confirm that LoRA fine-tuning significantly enhances LLM performance, surpassing non-fine-tuned bases and GPT-4. The standout performance of models like Mistral-7B across multiple tasks highlights the importance of base model selection in fine-tuning success. We find that dataset complexity heuristics can be reasonably leveraged as potential predictors of fine-tuning success, suggesting that the nature of the task plays an important role in the effectiveness of fine-tuning.

Despite these outcomes, limitations such as the scale of evaluations, training constraints, and the simplicity of our prompt engineering approaches suggest areas for future improvement. We release all of our models and training setups for further community validation and experimentation.

On serving, we demonstrate the practical deployment of these models using the LoRAX framework through the LoRA Land web application. We provide benchmarks for time to first token (TFTT), total request time, and token streaming time, and measure LoRAX's latency robustness to up to 100 concurrent users. 

Altogether, LoRA Land emphasizes the quality and cost-effectiveness of employing multiple specialized LLMs over a single, general-purpose LLM.

\section{Acknowledgements}
\label{sec:contrib}

Justin Zhao led the research and wrote the paper. Justin Zhao and Timothy Wang designed the experiments, created the evaluation harness, ran experiments, and analyzed the data. Wael Abid led LoRAX performance benchmarks and wrote section 6 of the paper. Piero Molino was an early advocate for the idea and provided feedback on the writing, experiments, and data analysis. We thank Martin Davis, Kabir Brar, and Jackie Ho for designing and developing the LoRA Land web application. We thank Travis Addair, Geoffrey Angus, Magdy Saleh, Noah Yoshida, Jeffrey Tang, and open source contributors for developing LoRAX. We thank Noah Yoshida and Gyanesh Mishra for supporting deployments. We thank Arnav Garg, Geoffrey Angus, Arnav Garg, Jeff Kinnison, Alex Shertinsky, Travis Addair, Piero Molino, and open source contributors for Ludwig. We thank Will Gorman, Michael Gonzales, and Devvret Rishi for support, discussion, and feedback.

\bibliographystyle{plainnat}
{
\small
\bibliography{anthology}
}
\clearpage
\appendix
\section{Prompts for all tasks}

\label{sec:appendixprompts}

The preprocessing code, prompts, configuration, and splits used for all experiments can be found at \href{https://github.com/predibase/lora_bakeoff}{https://github.com/predibase/lora\_bakeoff}. 

\section{LoRAX benchmarking scripts}

\label{sec:appendixscripts}

The load testing script and instructions can be found at \href{https://github.com/predibase/lora_bakeoff}{https://github.com/predibase/lora\_bakeoff}. 

\begin{landscape}

\section{Full Results Tables}
\label{sec:appendixresults}

\begin{table}[ht]
\centering

\scalebox{0.55}{

\begin{tabular}{ccccccccccccccc}

\multirow{2}{*}{\textbf{Category}} & \multirow{2}{*}{\textbf{Task}} & \multirow{2}{*}{\textbf{Metric}} & \textbf{Microsoft} & \multicolumn{4}{c}{\textbf{Google}}                                                             & \multicolumn{2}{c}{\textbf{Meta}}              & \multicolumn{2}{c}{\textbf{Mistral}}               & \textbf{Hugging Face}   & \multicolumn{2}{c}{\textbf{OpenAI}}     \\
                                   &                                &                                  & \textbf{phi-2}     & \textbf{gemma-2b} & \textbf{gemma-2b-instruct} & \textbf{gemma-7b} & \textbf{gemma-7b-instruct} & \textbf{llama-2-7b} & \textbf{llama-2-7b-chat} & \textbf{mistral-7b} & \textbf{mistral-7b-instruct} & \textbf{zephyr-7b-beta} & \textbf{gpt-3.5-turbo} & \textbf{gpt-4} \\
\midrule
\multirow{7}{*}{Classic NLP}       & bc5cdr                         & rouge                            & 0.172              & 0.013             & 0.494                      & 0.075             & 0.198                      & 0.185               & 0.024                    & 0.177               & 0.703                        & 0.146                   & 0.732                  & \textbf{0.890} \\
                                   & conllpp                        & rouge                            & 0.101              & 0.011             & 0.647                      & 0.085             & 0.120                      & 0.108               & 0.115                    & 0.148               & 0.733                        & 0.088                   & \textbf{0.810}         & 0.742          \\
                                   & e2e\_nlg                       & rouge                            & 0.132              & 0.174             & 0.281                      & 0.152             & 0.434                      & 0.087               & 0.442                    & 0.167               & 0.482                        & 0.122                   & 0.467                  & \textbf{0.513} \\
                                   & tldr\_content\_gen             & rouge                            & 0.158              & 0.117             & 0.160                      & 0.089             & 0.141                      & 0.148               & 0.183                    & 0.153               & 0.163                        & 0.164                   & 0.173                  & 0.125          \\
                                   & tldr\_headline\_gen            & rouge                            & 0.169              & 0.034             & 0.155                      & 0.063             & 0.152                      & 0.078               & 0.174                    & 0.071               & 0.171                        & 0.120                   & \textbf{0.195}         & 0.175          \\
                                   & viggo                          & rouge                            & 0.133              & 0.093             & 0.237                      & 0.123             & 0.313                      & 0.141               & 0.356                    & 0.044               & 0.374                        & 0.193                   & 0.372                  & \textbf{0.374} \\
                                   & webnlg                         & rouge                            & 0.120              & 0.055             & 0.312                      & 0.257             & 0.453                      & 0.148               & 0.563                    & 0.091               & 0.541                        & 0.512                   & 0.581                  & \textbf{0.583} \\
\midrule
\multirow{2}{*}{Coding}            & magicoder                      & humaneval                        & 0.012              & 0.037             & 0.024                      & 0.030             & 0.018                      & 0.012               & 0.134                    & 0.201               & 0.152                        & 0.049                   & 0.683                  & \textbf{0.829} \\
                                   & wikisql                        & rouge                            & 0.143              & 0.030             & 0.301                      & 0.036             & 0.244                      & 0.043               & 0.093                    & 0.265               & 0.134                        & 0.080                   & 0.887                  & \textbf{0.909} \\
\midrule
\multirow{8}{*}{Knowledge}         & boolq                          & accuracy                         & 0.691              & 0.447             & 0.661                      & 0.300             & 0.735                      & 0.645               & 0.759                    & 0.669               & 0.764                        & 0.683                   & 0.870                  & \textbf{0.911} \\
                                   & dbpedia                        & dbpedia                          & 0.268              & 0.018             & 0.086                      & 0.021             & 0.089                      & 0.043               & 0.868                    & 0.036               & 0.313                        & 0.578                   & 0.853                  & \textbf{0.965} \\
                                   & customer\_support              & accuracy                         & 0.250              & 0.120             & 0.380                      & 0.100             & 0.850                      & 0.110               & 0.630                    & 0.030               & 0.730                        & 0.540                   & \textbf{1.000}         & \textbf{1.000} \\
                                   & glue\_qnli                     & accuracy                         & 0.496              & 0.439             & 0.444                      & 0.463             & 0.685                      & 0.510               & 0.736                    & 0.533               & 0.743                        & 0.569                   & 0.829                  & \textbf{0.902} \\
                                   & glue\_stsb                     & mae                              & 0.682              & 0.197             & 0.590                      & 0.537             & 0.729                      & 0.651               & 0.680                    & 0.672               & 0.723                        & 0.814                   & \textbf{0.857}         & 0.773          \\
                                   & legal                          & rouge                            & 0.008              & 0.010             & 0.037                      & 0.019             & 0.053                      & 0.009               & 0.026                    & 0.001               & 0.158                        & 0.039                   & 0.266                  & \textbf{0.305} \\
                                   & reuters                        & rouge                            & 0.003              & 0.001             & 0.010                      & 0.001             & 0.009                      & 0.003               & 0.010                    & 0.004               & 0.010                        & 0.005                   & \textbf{0.026}         & 0.014          \\
                                   & mmlu                           & accuracy                         & 0.339              & 0.160             & 0.279                      & 0.302             & 0.460                      & 0.189               & 0.349                    & 0.402               & 0.446                        & 0.506                   & 0.504                  & \textbf{0.774} \\
\midrule
\multirow{13}{*}{Reasoning}        & winogrande                     & accuracy                         & 0.380              & 0.309             & 0.515                      & 0.390             & 0.576                      & 0.503               & 0.515                    & 0.498               & 0.546                        & 0.532                   & 0.569                  & \textbf{0.832} \\
                                   & arc\_combined                  & accuracy                         & 0.323              & 0.180             & 0.254                      & 0.272             & 0.657                      & 0.304               & 0.379                    & 0.573               & 0.673                        & 0.497                   & 0.926                  & \textbf{0.947} \\
                                   & glue\_cola                     & accuracy                         & 0.463              & 0.152             & 0.642                      & 0.062             & 0.749                      & 0.691               & 0.691                    & 0.691               & 0.797                        & 0.788                   & 0.843                  & \textbf{0.864} \\
                                   & glue\_mnli                     & accuracy                         & 0.328              & 0.053             & 0.347                      & 0.213             & 0.272                      & 0.315               & 0.293                    & 0.327               & 0.455                        & 0.348                   & 0.588                  & \textbf{0.803} \\
                                   & glue\_mrpc                     & accuracy                         & 0.652              & 0.265             & 0.664                      & 0.654             & 0.652                      & 0.679               & 0.674                    & 0.684               & 0.694                        & 0.676                   & 0.689                  & \textbf{0.777} \\
                                   & glue\_qqp                      & accuracy                         & 0.327              & 0.138             & 0.337                      & 0.316             & 0.396                      & 0.345               & 0.340                    & 0.327               & 0.708                        & 0.340                   & 0.830                  & \textbf{0.841} \\
                                   & glue\_sst2                     & accuracy                         & 0.487              & 0.407             & 0.719                      & 0.187             & 0.682                      & 0.306               & 0.695                    & 0.115               & 0.933                        & 0.706                   & 0.933                  & \textbf{0.942} \\
                                   & glue\_wnli                     & accuracy                         & 0.437              & 0.183             & 0.437                      & 0.366             & 0.437                      & 0.423               & 0.437                    & 0.437               & 0.437                        & 0.437                   & 0.521                  & \textbf{0.930} \\
                                   & covid                          & accuracy                         & 0.207              & 0.154             & 0.317                      & 0.169             & 0.322                      & 0.162               & 0.212                    & 0.191               & 0.297                        & 0.243                   & \textbf{0.334}         & 0.309          \\
                                   & hellaswag                      & accuracy                         & 0.371              & 0.117             & 0.023                      & 0.112             & 0.201                      & 0.381               & 0.264                    & 0.246               & 0.249                        & 0.393                   & 0.622                  & \textbf{0.805} \\
                                   & hellaswag\_processed           & rouge                            & 0.037              & 0.056             & \textbf{0.146}             & 0.109             & 0.143                      & 0.044               & 0.089                    & 0.038               & 0.134                        & 0.040                   & 0.140                  & 0.134          \\
                                   & jigsaw                         & accuracy                         & 0.491              & 0.490             & 0.482                      & 0.233             & 0.520                      & 0.486               & 0.545                    & 0.475               & 0.704                        & 0.472                   & 0.735                  & \textbf{0.754} \\
                                   & drop                           & rouge                            & 0.018              & 0.013             & 0.034                      & 0.024             & 0.042                      & 0.010               & 0.047                    & 0.011               & 0.066                        & 0.023                   & 0.119                  & \textbf{0.393} \\
\midrule
Math                               & gsm8k                          & accuracy                         & 0.083              & 0.026             & 0.082                      & 0.039             & 0.364                      & 0.051               & 0.160                    & 0.114               & 0.275                        & 0.133                   & \textbf{0.622}         & 0.373         
\end{tabular}
}

\label{table:basemodelsperformance}
\vspace{10mm}
\caption{Base model performance for every task and base model, before fine-tuning.}

\end{table}

\begin{table}[ht]
\centering

\scalebox{0.55}{
\begin{tabular}{ccccccccccccccc}
\multirow{2}{*}{\textbf{Category}} & \multirow{2}{*}{\textbf{Task}} & \multirow{2}{*}{\textbf{Metric}} & \textbf{Microsoft} & \multicolumn{4}{c}{\textbf{Google}}                                                             & \multicolumn{2}{c}{\textbf{Meta}}              & \multicolumn{2}{c}{\textbf{Mistral}}               & \textbf{Hugging Face}   & \multicolumn{2}{c}{\textbf{OpenAI}}     \\
                                   &                                &                                  & \textbf{phi-2}     & \textbf{gemma-2b} & \textbf{gemma-2b-instruct} & \textbf{gemma-7b} & \textbf{gemma-7b-instruct} & \textbf{llama-2-7b} & \textbf{llama-2-7b-chat} & \textbf{mistral-7b} & \textbf{mistral-7b-instruct} & \textbf{zephyr-7b-beta} & \textbf{gpt-3.5-turbo} & \textbf{gpt-4} \\
\midrule
\multirow{7}{*}{Classic NLP}       & bc5cdr                         & rouge                            & 0.950 (+0.778)     & 0.961 (+0.948)    & 0.956 (+0.462)             & 0.969 (+0.894)    & 0.969 (+0.771)             & 0.967 (+0.782)      & 0.967 (+0.943)           & 0.972 (+0.795)      & 0.971 (+0.268)               & 0.969 (+0.823)          & 0.732                  & 0.890          \\
                                   & conllpp                        & rouge                            & 0.950 (+0.849)     & 0.976 (+0.965)    & 0.975 (+0.328)             & 0.989 (+0.904)    & 0.989 (+0.869)             & 0.977 (+0.869)      & 0.980 (+0.865)           & 0.986 (+0.838)      & 0.987 (+0.254)               & 0.984 (+0.896)          & 0.810                  & 0.742          \\
                                   & e2e\_nlg                       & rouge                            & 0.516 (+0.384)     & 0.543 (+0.369)    & 0.543 (+0.262)             & 0.549 (+0.397)    & 0.550 (+0.116)             & 0.541 (+0.454)      & 0.538 (+0.096)           & 0.552 (+0.385)      & 0.551 (+0.069)               & 0.543 (+0.421)          & 0.467                  & 0.513          \\
                                   & tldr\_content\_gen             & rouge                            & 0.201 (+0.043)     & 0.204 (+0.087)    & 0.202 (+0.042)             & 0.217 (+0.128)    & 0.194 (+0.053)             & 0.219 (+0.071)      & 0.220 (+0.037)           & 0.227 (+0.074)      & 0.226 (+0.063)               & 0.230 (+0.066)          & 0.173                  & 0.125          \\
                                   & tldr\_headline\_gen            & rouge                            & 0.343 (+0.174)     & 0.404 (+0.370)    & 0.385 (+0.230)             & 0.394 (+0.331)    & 0.391 (+0.239)             & 0.432 (+0.354)      & 0.429 (+0.255)           & 0.434 (+0.363)      & 0.419 (+0.248)               & 0.441 (+0.321)          & 0.195                  & 0.175          \\
                                   & viggo                          & rouge                            & 0.445 (+0.312)     & 0.504 (+0.411)    & 0.497 (+0.260)             & 0.474 (+0.351)    & 0.441 (+0.128)             & 0.469 (+0.328)      & 0.463 (+0.107)           & 0.483 (+0.439)      & 0.505 (+0.131)               & 0.477 (+0.284)          & 0.372                  & 0.374          \\
                                   & webnlg                         & rouge                            & 0.634 (+0.514)     & 0.652 (+0.597)    & 0.649 (+0.337)             & 0.673 (+0.416)    & 0.664 (+0.211)             & 0.666 (+0.518)      & 0.673 (+0.110)           & 0.681 (+0.590)      & 0.672 (+0.131)               & 0.677 (+0.165)          & 0.581                  & 0.583          \\
\midrule
\multirow{2}{*}{Coding}            & magicoder                      & humaneval                        & 0.384 (+0.372)     & 0.079 (+0.042)    & 0.152 (+0.128)             & 0.433 (+0.403)    & 0.329 (+0.311)             & 0.122 (+0.110)      & 0.152 (+0.018)           & 0.335 (+0.134)      & 0.341 (+0.189)               & 0.317 (+0.268)          & 0.683                  & 0.829          \\
                                   & wikisql                        & rouge                            & 0.680 (+0.537)     & 0.890 (+0.860)    & 0.885 (+0.584)             & 0.894 (+0.858)    & 0.893 (+0.649)             & 0.898 (+0.855)      & 0.893 (+0.800)           & 0.669 (+0.404)      & 0.651 (+0.517)               & 0.896 (+0.816)          & 0.887                  & 0.909          \\
\midrule
\multirow{8}{*}{Knowledge}         & boolq                          & accuracy                         & 0.863 (+0.172)     & 0.811 (+0.364)    & 0.776 (+0.115)             & 0.664 (+0.364)    & 0.665 (-0.070)             & 0.884 (+0.239)      & 0.872 (+0.113)           & 0.909 (+0.240)      & 0.891 (+0.127)               & 0.897 (+0.214)          & 0.870                  & 0.911          \\
                                   & dbpedia                        & accuracy                         & 0.988 (+0.720)     & 0.960 (+0.942)    & 0.961 (+0.875)             & 0.964 (+0.943)    & 0.971 (+0.882)             & 0.975 (+0.932)      & 0.980 (+0.112)           & 0.981 (+0.945)      & 0.970 (+0.657)               & 0.963 (+0.385)          & 0.853                  & 0.965          \\
                                   & customer\_support              & accuracy                         & 1.000 (+0.750)     & 1.000 (+0.880)    & 1.000 (+0.620)             & 1.000 (+0.900)    & 1.000 (+0.150)             & 1.000 (+0.890)      & 1.000 (+0.370)           & 1.000 (+0.970)      & 1.000 (+0.270)               & 1.000 (+0.460)          & 1.000                  & 1.000          \\
                                   & glue\_qnli                     & accuracy                         & 0.892 (+0.396)     & 0.872 (+0.433)    & 0.887 (+0.443)             & 0.897 (+0.434)    & 0.876 (+0.191)             & 0.860 (+0.350)      & 0.925 (+0.189)           & 0.931 (+0.398)      & 0.906 (+0.163)               & 0.928 (+0.359)          & 0.829                  & 0.902          \\
                                   & glue\_stsb                     & mae                              & 0.888 (+0.206)     & 0.875 (+0.678)    & 0.895 (+0.305)             & 0.704 (+0.167)    & 0.893 (+0.164)             & 0.912 (+0.261)      & 0.907 (+0.227)           & 0.913 (+0.241)      & 0.911 (+0.188)               & 0.911 (+0.097)          & 0.857                  & 0.773          \\
                                   & legal                          & rouge                            & 0.404 (+0.396)     & 0.503 (+0.493)    & 0.451 (+0.414)             & 0.586 (+0.567)    & 0.580 (+0.527)             & 0.668 (+0.659)      & 0.602 (+0.576)           & 0.602 (+0.601)      & 0.666 (+0.508)               & 0.683 (+0.644)          & 0.266                  & 0.305          \\
                                   & reuters                        & rouge                            & 0.149 (+0.146)     & 0.458 (+0.457)    & 0.465 (+0.455)             & 0.475 (+0.474)    & 0.477 (+0.468)             & 0.475 (+0.472)      & 0.475 (+0.465)           & 0.431 (+0.427)      & 0.455 (+0.445)               & 0.479 (+0.474)          & 0.026                  & 0.014          \\
                                   & mmlu                           & accuracy                         & 0.530 (+0.191)     & 0.446 (+0.286)    & 0.432 (+0.153)             & 0.248 (-0.054)    & 0.243 (-0.217)             & 0.519 (+0.330)      & 0.526 (+0.177)           & 0.561 (+0.159)      & 0.558 (+0.112)               & 0.589 (+0.083)          & 0.504                  & 0.774          \\
\midrule
\multirow{13}{*}{Reasoning}        & winogrande                     & accuracy                         & 0.741 (+0.361)     & 0.493 (+0.184)    & 0.494 (-0.021)             & 0.493 (+0.103)    & 0.493 (-0.083)             & 0.493 (-0.010)      & 0.754 (+0.239)           & 0.840 (+0.342)      & 0.818 (+0.272)               & 0.825 (+0.293)          & 0.569                  & 0.832          \\
                                   & arc\_combined                  & accuracy                         & 0.915 (+0.592)     & 0.768 (+0.588)    & 0.745 (+0.491)             & 0.269 (-0.003)    & 0.258 (-0.399)             & 0.832 (+0.528)      & 0.843 (+0.464)           & 0.915 (+0.342)      & 0.857 (+0.184)               & 0.909 (+0.412)          & 0.926                  & 0.947          \\
                                   & glue\_cola                     & accuracy                         & 0.843 (+0.380)     & 0.828 (+0.676)    & 0.777 (+0.135)             & 0.691 (+0.629)    & 0.691 (-0.058)             & 0.837 (+0.146)      & 0.860 (+0.169)           & 0.845 (+0.154)      & 0.849 (+0.052)               & 0.872 (+0.084)          & 0.843                  & 0.864          \\
                                   & glue\_mnli                     & accuracy                         & 0.871 (+0.543)     & 0.833 (+0.780)    & 0.837 (+0.490)             & 0.882 (+0.669)    & 0.874 (+0.602)             & 0.877 (+0.562)      & 0.870 (+0.577)           & 0.893 (+0.566)      & 0.887 (+0.432)               & 0.899 (+0.551)          & 0.588                  & 0.803          \\
                                   & glue\_mrpc                     & accuracy                         & 0.858 (+0.206)     & 0.850 (+0.585)    & 0.870 (+0.206)             & 0.740 (+0.086)    & 0.684 (+0.032)             & 0.797 (+0.118)      & 0.870 (+0.196)           & 0.887 (+0.203)      & 0.885 (+0.191)               & 0.870 (+0.194)          & 0.689                  & 0.777          \\
                                   & glue\_qqp                      & accuracy                         & 0.875 (+0.548)     & 0.877 (+0.739)    & 0.863 (+0.526)             & 0.872 (+0.556)    & 0.673 (+0.277)             & 0.868 (+0.523)      & 0.874 (+0.534)           & 0.870 (+0.543)      & 0.883 (+0.175)               & 0.867 (+0.527)          & 0.830                  & 0.841          \\
                                   & glue\_sst2                     & accuracy                         & 0.946 (+0.459)     & 0.954 (+0.547)    & 0.919 (+0.200)             & 0.919 (+0.732)    & 0.943 (+0.261)             & 0.948 (+0.642)      & 0.956 (+0.261)           & 0.959 (+0.844)      & 0.958 (+0.025)               & 0.961 (+0.255)          & 0.933                  & 0.942          \\
                                   & glue\_wnli                     & accuracy                         & 0.676 (+0.239)     & 0.563 (+0.380)    & 0.563 (+0.126)             & 0.563 (+0.197)    & 0.563 (+0.126)             & 0.718 (+0.295)      & 0.775 (+0.338)           & 0.873 (+0.436)      & 0.803 (+0.366)               & 0.831 (+0.394)          & 0.521                  & 0.930          \\
                                   & covid                          & accuracy                         & 0.692 (+0.485)     & 0.827 (+0.673)    & 0.832 (+0.515)             & 0.830 (+0.661)    & 0.843 (+0.521)             & 0.751 (+0.589)      & 0.727 (+0.515)           & 0.770 (+0.579)      & 0.811 (+0.514)               & 0.776 (+0.533)          & 0.334                  & 0.309          \\
                                   & hellaswag                      & accuracy                         & 0.714 (+0.343)     & 0.397 (+0.280)    & 0.252 (+0.229)             & 0.252 (+0.140)    & 0.252 (+0.051)             & 0.741 (+0.360)      & 0.736 (+0.472)           & 0.834 (+0.588)      & 0.730 (+0.481)               & 0.828 (+0.435)          & 0.622                  & 0.805          \\
                                   & hellaswag\_processed           & rouge                            & 0.223 (+0.186)     & 0.235 (+0.179)    & 0.214 (+0.068)             & 0.222 (+0.113)    & 0.208 (+0.065)             & 0.253 (+0.209)      & 0.249 (+0.160)           & 0.261 (+0.223)      & 0.254 (+0.120)               & 0.260 (+0.220)          & 0.140                  & 0.134          \\
                                   & jigsaw                         & accuracy                         & 0.824 (+0.333)     & 0.852 (+0.362)    & 0.845 (+0.363)             & 0.824 (+0.591)    & 0.789 (+0.269)             & 0.847 (+0.361)      & 0.832 (+0.287)           & 0.849 (+0.374)      & 0.867 (+0.163)               & 0.866 (+0.394)          & 0.735                  & 0.754          \\
                                   & drop                           & rouge                            & 0.549 (+0.531)     & 0.506 (+0.493)    & 0.410 (+0.376)             & 0.693 (+0.669)    & 0.602 (+0.560)             & 0.670 (+0.660)      & 0.667 (+0.620)           & 0.705 (+0.694)      & 0.677 (+0.611)               & 0.741 (+0.718)          & 0.119                  & 0.393          \\
\midrule
Math                               & gsm8k                          & accuracy                         & 0.441 (+0.358)     & 0.258 (+0.232)    & 0.240 (+0.158)             & 0.569 (+0.530)    & 0.505 (+0.141)             & 0.339 (+0.288)      & 0.323 (+0.163)           & 0.520 (+0.406)      & 0.488 (+0.213)               & 0.503 (+0.370)          & 0.622                  & 0.373         
\end{tabular}
}

\label{table:finetunedmodelsperformance}
\vspace{10mm}
\caption{Performance of 310 fine-tuned models across 10 base models and 31 tasks. The value in parentheses is the absolute improvement compared to the base model. Fine-tuning scores were not obtained for GPT-3.5-Turbo or GPT-4.}

\end{table}

\begin{table}[ht]

\centering

\scalebox{0.6}{

\begin{tabular}{ccccccccccccccccccccc}
\textbf{Task}        & \textbf{\begin{tabular}[c]{@{}c@{}}Max\\ GPT-4 \\ Lift\end{tabular}} & \textbf{\begin{tabular}[c]{@{}c@{}}Average\\ Base\\ Model\\ Lift\end{tabular}} & \textbf{\begin{tabular}[c]{@{}c@{}}Best\\ Base\\ Model\\ Score\end{tabular}} & \textbf{\begin{tabular}[c]{@{}c@{}}Average\\ Base\\ Model\\ Score\end{tabular}} & \textbf{\begin{tabular}[c]{@{}c@{}}Best\\ Fine-tuned\\ Score\end{tabular}} & \textbf{\begin{tabular}[c]{@{}c@{}}Average\\ Fine-Tuned\\ Score\end{tabular}} & \textbf{\begin{tabular}[c]{@{}c@{}}Input\\ length \\ p95\end{tabular}} & \textbf{\begin{tabular}[c]{@{}c@{}}Input\\ length\\ $\mu$\end{tabular}} & \textbf{\begin{tabular}[c]{@{}c@{}}Input\\ length\\ $\sigma$\end{tabular}} & \textbf{\begin{tabular}[c]{@{}c@{}}Output\\ length\\ p95\end{tabular}} & \textbf{\begin{tabular}[c]{@{}c@{}}Output\\ length\\ $\mu$\end{tabular}} & \textbf{\begin{tabular}[c]{@{}c@{}}Output\\ length\\ $\sigma$\end{tabular}} & \textbf{\begin{tabular}[c]{@{}c@{}}Example\\ length\\ $\mu$\end{tabular}} & \textbf{\begin{tabular}[c]{@{}c@{}}Example\\ length\\ p95\end{tabular}} & \textbf{\begin{tabular}[c]{@{}c@{}}Example\\ length\\ $\sigma$\end{tabular}} & \textbf{\begin{tabular}[c]{@{}c@{}}I/O\\ rougeL\\ similarity\\ $\mu$\end{tabular}} & \textbf{\begin{tabular}[c]{@{}c@{}}I/O\\ rougeL\\ similarity\\ $\sigma$\end{tabular}} & \textbf{\begin{tabular}[c]{@{}c@{}}Compr.\\ $\mu$\end{tabular}} & \textbf{\begin{tabular}[c]{@{}c@{}}Compr.\\ $\sigma$\end{tabular}} & \textbf{\begin{tabular}[c]{@{}c@{}}\#\\ training\\ examples\end{tabular}} \\
\midrule
arc\_combined        & -0.032                                                               & 0.320                                                                          & 0.673                                                                        & 0.411                                                                           & 0.915                                                                      & 0.731                                                                         & 143                                                                    & 102.89                                                              & 21.68                                                               & 1                                                                      & 1.00                                                                 & 0.00                                                                 & 102.92                                                                & 143.00                                                                  & 21.659                                                                & 0.034                                                                          & 0.009                                                                          & 0.644                                                       & 0.064                                                       & 3370                                                                      \\
bc5cdr               & 0.082                                                                & 0.746                                                                          & 0.703                                                                        & 0.219                                                                           & 0.972                                                                      & 0.965                                                                         & 175                                                                    & 142.15                                                              & 19.17                                                               & 58                                                                     & 37.11                                                                & 11.27                                                                & 178.26                                                                & 226.05                                                                  & 27.839                                                                & 0.191                                                                          & 0.026                                                                          & 0.547                                                       & 0.014                                                       & 5228                                                                      \\
boolq                & -0.002                                                               & 0.188                                                                          & 0.764                                                                        & \textbf{0.635}                                                                  & 0.909                                                                      & 0.823                                                                         & 270.7                                                                  & 145.23                                                              & 69.03                                                               & 1                                                                      & 1.00                                                                 & 0.00                                                                 & 146.23                                                                & 271.70                                                                  & 69.031                                                                & 0.000                                                                          & 0.003                                                                          & 0.596                                                       & 0.066                                                       & 9427                                                                      \\
conllpp              & 0.247                                                                & 0.764                                                                          & 0.733                                                                        & 0.216                                                                           & 0.989                                                                      & 0.979                                                                         & 137                                                                    & 111.88                                                              & 13.17                                                               & 38                                                                     & 24.88                                                                & 7.58                                                                 & 135.76                                                                & 170.00                                                                  & 18.647                                                                & 0.126                                                                          & 0.031                                                                          & 0.583                                                       & 0.013                                                       & 14041                                                                     \\
covid                & 0.534                                                                & 0.559                                                                          & 0.322                                                                        & 0.227                                                                           & 0.843                                                                      & 0.786                                                                         & 222                                                                    & 189.89                                                              & 19.85                                                               & 3                                                                      & 1.58                                                                 & 0.91                                                                 & 190.18                                                                & 223.00                                                                  & 19.910                                                                & 0.020                                                                          & 0.007                                                                          & 0.570                                                       & 0.012                                                       & 37361                                                                     \\
customer\_support    & 0.000                                                                & 0.626                                                                          & 0.850                                                                        & 0.374                                                                           & 1.000                                                                      & 1.000                                                                         & 376                                                                    & 274.02                                                              & 57.26                                                               & 3                                                                      & 2.13                                                                 & 0.34                                                                 & 275.15                                                                & 377.00                                                                  & 57.160                                                                & 0.023                                                                          & 0.007                                                                          & 0.472                                                       & 0.034                                                       & 1245                                                                      \\
dbpedia              & 0.023                                                                & 0.739                                                                          & 0.868                                                                        & 0.232                                                                           & 0.988                                                                      & 0.971                                                                         & 210                                                                    & 162.20                                                              & 30.93                                                               & 4                                                                      & 1.77                                                                 & 1.00                                                                 & 162.83                                                                & 211.00                                                                  & 31.021                                                                & 0.023                                                                          & 0.006                                                                          & 0.617                                                       & 0.030                                                       & 560000                                                                    \\
drop                 & 0.348                                                                & 0.593                                                                          & 0.066                                                                        & 0.029                                                                           & 0.741                                                                      & 0.622                                                                         & 570                                                                    & 335.17                                                              & 150.52                                                              & 5                                                                      & 2.05                                                                 & 1.58                                                                 & 337.16                                                                & 571.00                                                                  & 150.431                                                               & 0.009                                                                          & 0.012                                                                          & 0.518                                                       & 0.039                                                       & 77400                                                                     \\
e2e\_nlg             & 0.039                                                                & 0.295                                                                          & 0.482                                                                        & 0.247                                                                           & 0.552                                                                      & 0.543                                                                         & 116                                                                    & 104.18                                                              & 7.38                                                                & 40                                                                     & 25.08                                                                & 8.33                                                                 & 128.12                                                                & 153.00                                                                  & 14.427                                                                & 0.173                                                                          & 0.050                                                                          & 0.513                                                       & 0.018                                                       & 42061                                                                     \\
glue\_cola           & 0.008                                                                & 0.237                                                                          & 0.797                                                                        & \textbf{0.573}                                                                  & 0.872                                                                      & 0.809                                                                         & 58                                                                     & 50.34                                                               & 4.08                                                                & 2                                                                      & 1.10                                                                 & 0.30                                                                 & 51.34                                                                 & 59.00                                                                   & 4.075                                                                 & 0.062                                                                          & 0.006                                                                          & 0.646                                                       & 0.010                                                       & 8551                                                                      \\
glue\_mnli           & 0.096                                                                & 0.577                                                                          & 0.455                                                                        & 0.295                                                                           & 0.899                                                                      & 0.872                                                                         & 127                                                                    & 94.73                                                               & 18.76                                                               & 1                                                                      & 1.00                                                                 & 0.00                                                                 & 95.73                                                                 & 128.00                                                                  & 18.763                                                                & 0.031                                                                          & 0.007                                                                          & 0.558                                                       & 0.023                                                       & 392702                                                                    \\
glue\_mrpc           & 0.110                                                                & 0.202                                                                          & 0.694                                                                        & 0.629                                                                           & 0.887                                                                      & 0.831                                                                         & 122                                                                    & 100.78                                                              & 13.18                                                               & 1                                                                      & 1.00                                                                 & 0.00                                                                 & 101.78                                                                & 123.00                                                                  & 13.179                                                                & 0.029                                                                          & 0.004                                                                          & 0.539                                                       & 0.038                                                       & 3668                                                                      \\
glue\_qnli           & 0.029                                                                & 0.336                                                                          & 0.743                                                                        & 0.562                                                                           & 0.931                                                                      & 0.897                                                                         & 122                                                                    & 88.49                                                               & 18.44                                                               & 1                                                                      & 1.04                                                                 & 0.20                                                                 & 89.49                                                                 & 123.00                                                                  & 18.444                                                                & 0.032                                                                          & 0.006                                                                          & 0.621                                                       & 0.030                                                       & 104743                                                                    \\
glue\_qqp            & 0.042                                                                & 0.495                                                                          & 0.708                                                                        & 0.357                                                                           & 0.883                                                                      & 0.852                                                                         & 101                                                                    & 77.35                                                               & 12.61                                                               & 2                                                                      & 1.49                                                                 & 0.50                                                                 & 78.35                                                                 & 102.00                                                                  & 12.612                                                                & 0.038                                                                          & 0.006                                                                          & 0.603                                                       & 0.030                                                       & 363846                                                                    \\
glue\_sst2           & 0.019                                                                & 0.423                                                                          & 0.933                                                                        & 0.524                                                                           & 0.961                                                                      & 0.946                                                                         & 62                                                                     & 42.33                                                               & 9.40                                                                & 1                                                                      & 1.03                                                                 & 0.16                                                                 & 43.33                                                                 & 63.00                                                                   & 9.403                                                                 & 0.059                                                                          & 0.011                                                                          & 0.652                                                       & 0.019                                                       & 67349                                                                     \\
glue\_stsb           & 0.140                                                                & 0.253                                                                          & 0.814                                                                        & 0.628                                                                           & 0.913                                                                      & 0.881                                                                         & 121                                                                    & 89.99                                                               & 13.95                                                               & 4                                                                      & 3.16                                                                 & 0.37                                                                 & 92.99                                                                 & 124.00                                                                  & 13.946                                                                & 0.038                                                                          & 0.025                                                                          & 0.576                                                       & 0.027                                                       & 5749                                                                      \\
glue\_wnli           & -0.057                                                               & 0.290                                                                          & 0.437                                                                        & \textbf{0.403}                                                                  & 0.873                                                                      & 0.693                                                                         & 133                                                                    & 96.20                                                               & 17.81                                                               & 2                                                                      & 1.17                                                                 & 0.38                                                                 & 97.20                                                                 & 134.00                                                                  & 17.809                                                                & 0.030                                                                          & 0.005                                                                          & 0.560                                                       & 0.031                                                       & 635                                                                       \\
gsm8k                & 0.196                                                                & 0.286                                                                          & 0.364                                                                        & 0.133                                                                           & 0.569                                                                      & 0.419                                                                         & 106                                                                    & 65.30                                                               & 21.13                                                               & 186                                                                    & 100.70                                                               & 43.79                                                                & 165.77                                                                & 276.00                                                                  & 57.679                                                                & 0.272                                                                          & 0.081                                                                          & 0.545                                                       & 0.073                                                       & 7473                                                                      \\
hellaswag            & 0.029                                                                & 0.338                                                                          & 0.393                                                                        & 0.236                                                                           & 0.834                                                                      & 0.574                                                                         & 339                                                                    & 253.99                                                              & 71.38                                                               & 3                                                                      & 2.66                                                                 & 0.75                                                                 & 256.48                                                                & 341.00                                                                  & 71.366                                                                & 0.009                                                                          & 0.006                                                                          & 0.524                                                       & 0.027                                                       & 39905                                                                     \\
hellaswag\_processed & 0.127                                                                & 0.154                                                                          & 0.146                                                                        & 0.084                                                                           & 0.261                                                                      & 0.238                                                                         & 142                                                                    & 111.15                                                              & 20.86                                                               & 56                                                                     & 30.85                                                                & 15.46                                                                & 140.97                                                                & 185.00                                                                  & 33.774                                                                & 0.111                                                                          & 0.040                                                                          & 0.564                                                       & 0.023                                                       & 39905                                                                     \\
jigsaw               & 0.113                                                                & 0.350                                                                          & 0.704                                                                        & 0.490                                                                           & 0.867                                                                      & 0.839                                                                         & 600                                                                    & 475.45                                                              & 58.46                                                               & 1                                                                      & 1.00                                                                 & 0.00                                                                 & 476.45                                                                & 601.00                                                                  & 58.457                                                                & 0.006                                                                          & 0.001                                                                          & 0.486                                                       & 0.006                                                       & 159571                                                                    \\
legal                & 0.378                                                                & 0.539                                                                          & 0.158                                                                        & 0.036                                                                           & 0.683                                                                      & 0.575                                                                         & 485.05                                                                 & 246.96                                                              & 107.88                                                              & 6                                                                      & 2.92                                                                 & 1.73                                                                 & 249.88                                                                & 489.00                                                                  & 107.919                                                               & 0.012                                                                          & 0.013                                                                          & 0.499                                                       & 0.040                                                       & 17000                                                                     \\
magicoder            & -0.396                                                               & 0.198                                                                          & 0.201                                                                        & \textbf{0.067}                                                                  & 0.433                                                                      & 0.264                                                                         & 473                                                                    & 305.39                                                              & 91.88                                                               & 436                                                                    & 231.40                                                               & 110.12                                                               & 535.80                                                                & 805.00                                                                  & 151.769                                                               & 0.253                                                                          & 0.089                                                                          & 0.366                                                       & 0.046                                                       & 75197                                                                     \\
mmlu                 & -0.185                                                               & 0.122                                                                          & 0.506                                                                        & 0.343                                                                           & 0.589                                                                      & 0.465                                                                         & 577                                                                    & 377.20                                                              & 153.00                                                              & 1                                                                      & 1.00                                                                 & 0.00                                                                 & 378.20                                                                & 578.00                                                                  & 152.998                                                               & 0.010                                                                          & 0.012                                                                          & 0.526                                                       & 0.076                                                       & 99842                                                                     \\
reuters              & 0.465                                                                & 0.428                                                                          & 0.010                                                                        & 0.006                                                                           & 0.479                                                                      & 0.434                                                                         & 635                                                                    & 239.80                                                              & 186.43                                                              & 8                                                                      & 2.99                                                                 & 3.18                                                                 & 242.24                                                                & 637.05                                                                  & 187.038                                                               & 0.003                                                                          & 0.008                                                                          & 0.625                                                       & 0.087                                                       & 13625                                                                     \\
tldr\_content\_gen   & 0.105                                                                & 0.066                                                                          & 0.183                                                                        & 0.148                                                                           & 0.230                                                                      & 0.214                                                                         & 53                                                                     & 44.38                                                               & 5.97                                                                & 159                                                                    & 95.09                                                                & 36.51                                                                & 138.33                                                                & 204.45                                                                  & 38.846                                                                & 0.128                                                                          & 0.040                                                                          & 0.576                                                       & 0.037                                                       & 7138                                                                      \\
tldr\_headline\_gen  & 0.266                                                                & 0.289                                                                          & 0.174                                                                        & 0.119                                                                           & 0.441                                                                      & 0.407                                                                         & 184                                                                    & 120.96                                                              & 36.50                                                               & 22                                                                     & 13.53                                                                & 5.98                                                                 & 133.34                                                                & 199.45                                                                  & 38.845                                                                & 0.086                                                                          & 0.050                                                                          & 0.588                                                       & 0.041                                                       & 7138                                                                      \\
viggo                & 0.131                                                                & 0.275                                                                          & 0.374                                                                        & 0.201                                                                           & 0.505                                                                      & 0.476                                                                         & 193                                                                    & 169.05                                                              & 13.10                                                               & 49                                                                     & 27.68                                                                & 11.54                                                                & 196.48                                                                & 240.00                                                                  & 23.486                                                                & 0.112                                                                          & 0.042                                                                          & 0.512                                                       & 0.016                                                       & 5103                                                                      \\
webnlg               & 0.098                                                                & 0.359                                                                          & 0.563                                                                        & \textbf{0.305}                                                                  & 0.681                                                                      & 0.664                                                                         & 176                                                                    & 125.11                                                              & 27.61                                                               & 51                                                                     & 20.85                                                                & 17.15                                                                & 145.67                                                                & 215.05                                                                  & 37.522                                                                & 0.129                                                                          & 0.092                                                                          & 0.530                                                       & 0.033                                                       & 13211                                                                     \\
wikisql              & -0.011                                                               & 0.688                                                                          & 0.301                                                                        & 0.137                                                                           & 0.898                                                                      & 0.825                                                                         & 1921                                                                   & 805.07                                                              & 1668.51                                                             & 26                                                                     & 15.60                                                                & 5.72                                                                 & 819.66                                                                & 1941.10                                                                 & 1669.119                                                              & 0.050                                                                          & 0.030                                                                          & 0.387                                                       & 0.080                                                       & 56355                                                                     \\
winogrande           & 0.008                                                                & 0.168                                                                          & 0.576                                                                        & 0.476                                                                           & 0.840                                                                      & 0.644                                                                         & 62                                                                     & 54.32                                                               & 4.02                                                                & 1                                                                      & 1.00                                                                 & 0.00                                                                 & 55.32                                                                 & 63                                                                      & 4.017                                                                 & 0.052                                                                          & 0.004                                                                          & 0.748                                                       & 0.024                                                       & 9248                                                                     
\end{tabular}
}

\label{table:taskcomplexityheuristics}
\vspace{10mm}
\caption{Task and Dataset complexity heuristics and model quality measurements, across all tasks.}

\end{table}

\end{landscape}

\section{Other}

\label{sec:appendixother}

\begin{figure}[ht]
    \centering
    \includegraphics[width=\textwidth]{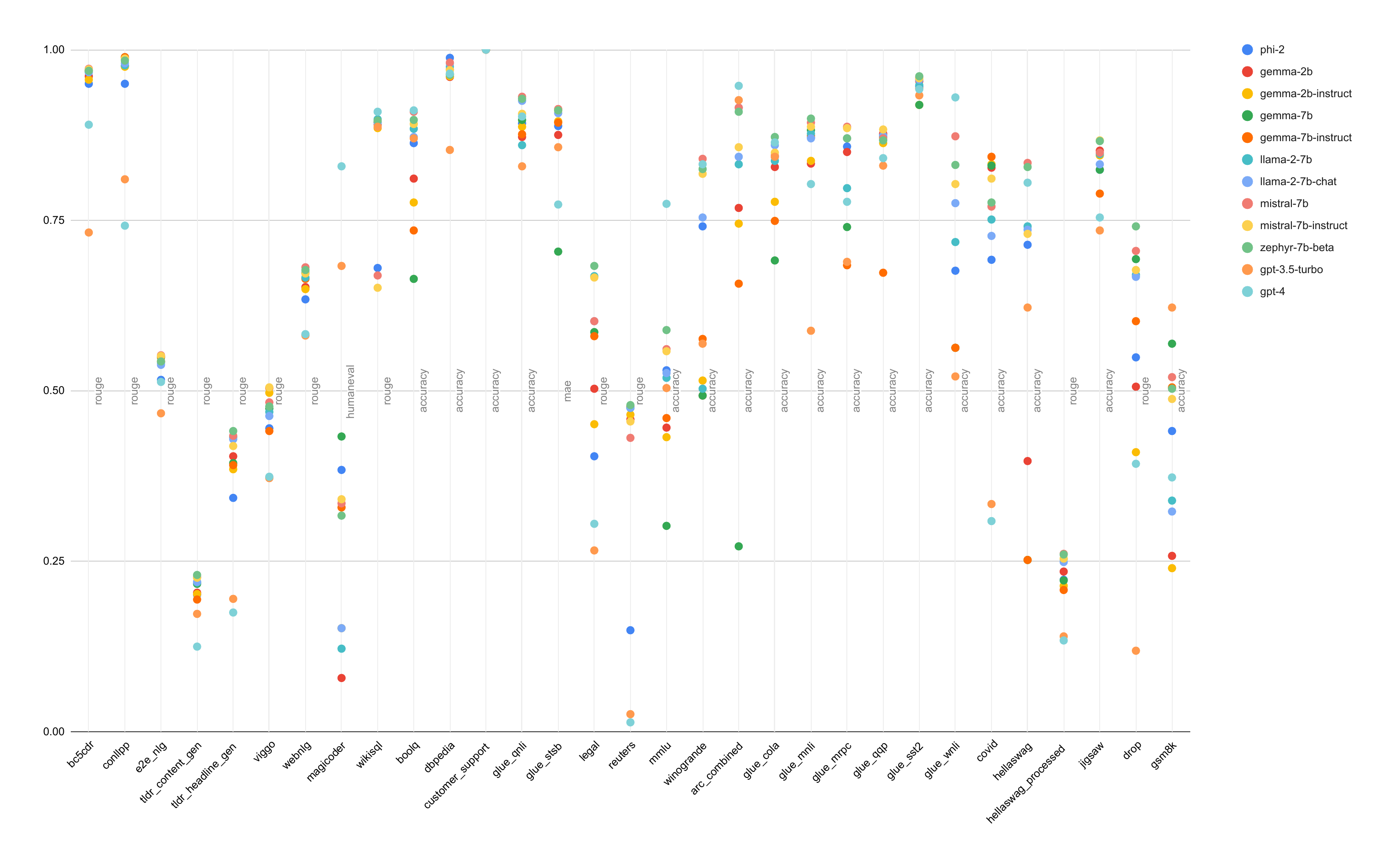}
    \caption{An esoteric visual representation of 310 fine-tuned LLMs.}
    \label{fig:art}
\end{figure}
\end{document}